\documentclass{article}

\usepackage{microtype}
\usepackage{graphicx}
\usepackage{subfigure}
\usepackage{booktabs} 

\usepackage{hyperref}

\usepackage[nonatbib, final]{neurips_2023}

\usepackage[utf8]{inputenc} % allow utf-8 input
\usepackage[T1]{fontenc}    % use 8-bit T1 fonts
\usepackage{hyperref}       % hyperlinks
\usepackage{url}            % simple URL typesetting
\usepackage{booktabs}       % professional-quality tables
\usepackage{amsfonts}       % blackboard math symbols
\usepackage{nicefrac}       % compact symbols for 1/2, etc.
\usepackage{microtype}      % microtypography
\usepackage{xcolor}         % colors

\usepackage{algorithm}
\usepackage{algorithmic}
\usepackage{xspace}
\usepackage{graphicx}
\usepackage{subfigure}
\usepackage{wrapfig}
\usepackage{graphicx}
\usepackage{subfigure}
\usepackage{url}
\usepackage{color}
\usepackage{multirow}
\usepackage{gensymb}
\usepackage{longtable}
\usepackage{tikz}
\usepackage{comment}
\usepackage{dsfont}
\usepackage{framed}
\usepackage{enumitem}
\usepackage{xspace}
\usepackage{bm}

\usepackage{multibib}
\newcites{newcite}{Reference}

\usepackage{amsthm}
\usepackage{amsmath}
\usepackage{amssymb}
\usepackage[capitalize,noabbrev]{cleveref}

\newtheorem{theorem}{Theorem}[section]
\newtheorem{corollary}{Corollary}[theorem]
\newtheorem{lemma}[theorem]{Lemma}
\newtheorem{proposition}[theorem]{Proposition}
\newtheorem{definition}{Definition}
\newtheorem{assumption}{Assumption}

\newcommand{\ouralg}{\texttt{RCL}\xspace}

\newcommand{\ouralgnt}{\texttt{RCL-O}\xspace}

\newcommand{\irobd}{iROBD\xspace}
\newcommand{\robd}{ROBD\xspace}
\newcommand{\ml}{ML\xspace}
\newcommand{\hitmin}{HitMin\xspace}
\newcommand{\ecltwoo}{EC-L2O\xspace}
\newcommand{\opt}{OPT\xspace}

\newcommand{\train}{robustification-aware\xspace}
\newcommand{\Train}{Robustification-aware\xspace}
\newcommand{\nontrain}{robustification-oblivious\xspace}
\newcommand{\Nontrain}{Robustification-oblivious\xspace}
\newcommand{\nontrainnoun}{robustification-obliviousness\xspace}

\title{Robust Learning for Smoothed Online Convex Optimization with Feedback Delay}

\author{%
Pengfei Li \\
University of California Riverside\\
Riverside, CA, USA\\
pli081@ucr.edu
\And
Jianyi Yang\\
University of California Riverside\\
Riverside, CA, USA\\
jyang239@ucr.edu
\And
Adam Wierman\\
California Institute of Technology\\
Pasadena, CA, USA\\
adamw@caltech.edu
\And 
Shaolei Ren\\
University of California Riverside\\
Riverside, CA, USA\\
shaolei@ucr.edu
}

\begin{document}

\maketitle

\begin{abstract}
We study a challenging form of Smoothed Online Convex Optimization, a.k.a. SOCO, including multi-step nonlinear switching costs and feedback delay. We propose a novel machine learning (ML) augmented online algorithm, Robustness-Constrained Learning
(\ouralg), which combines untrusted ML predictions with a trusted expert online algorithm via  constrained projection   to robustify the ML prediction.  Specifically,
we prove that \ouralg is able to guarantee
$(1+\lambda)$-competitiveness against any given expert for any
$\lambda>0$, while also explicitly training the ML model in a \train manner to improve the average-case performance.  Importantly,
\ouralg is the first ML-augmented algorithm with a provable robustness guarantee in the case of multi-step switching cost and feedback delay.
We demonstrate the improvement of \ouralg in both robustness and average performance using battery management for electrifying transportation
as a case study.
\end{abstract}

\section{Introduction}\label{sec:introduction}
This paper studies
\emph{Smoothed Online Convex Optimization (SOCO)}, 
a model that has seen application in a wide variety of settings.
The goal of SOCO is to minimize the sum of a per-round hitting cost and a switching cost that penalizes temporal changes in actions.  
The added (even single-step) switching cost creates substantial algorithmic challenges,
and has received more than a decade of attention 
 (see \cite{SOCO_MetricUntrustedPrediction_Google_ICML_2020_pmlr-v119-antoniadis20a,SOCO_DynamicRightSizing_Adam_Infocom_2011_LinWiermanAndrewThereska,SOCO_CapacityScalingAdaptiveBalancedGradient_Gatech_MAMA_2021_Sigmetrics_2021_10.1145/3512798.3512808,SOCO_Prediction_LinearSystem_NaLi_Harvard_NIPS_2019_10.5555/3454287.3455620,SOCO_Revisiting_Nanjing_NIPS_2021_zhang2021revisiting,SOCO_Memory_Adam_NIPS_2020_NEURIPS2020_ed46558a,SOCO_ML_ChasingConvexBodiesFunction_Adam_COLT_2022,SOCO_NonConvex_Adam_Sigmetrics_2020_10.1145/3379484,SOCO_OBD_Niangjun_Adam_COLT_2018_DBLP:conf/colt/ChenGW18,SOCO_Prediction_Error_Meta_ZhenhuaLiu_SIGMETRICS_2019_10.1145/3322205.3311087,SOCO_Prediction_RegularizedLookAhead_LeiJiao_XiaojunLin_INFOCOM_2021_9488766,Shaolei_L2O_ExpertCalibrated_SOCO_SIGMETRICS_Journal_2022} and the references therein).
While there have been various competitive online algorithms, e.g., ROBD,
to guarantee the worst-case performance robustness
for SOCO
 \cite{SOCO_OBD_R-OBD_Goel_Adam_NIPS_2019_NEURIPS2019_9f36407e,SOCO_Revisiting_Nanjing_NIPS_2021_zhang2021revisiting,SOCO_DynamicRightSizing_Adam_Infocom_2011_LinWiermanAndrewThereska,SOCO_Memory_Adam_NIPS_2020_NEURIPS2020_ed46558a,SOCO_NonConvex_Adam_Sigmetrics_2020_10.1145/3379484,SOCO_OBD_R-OBD_Goel_Adam_NIPS_2019_NEURIPS2019_9f36407e}, their average performance is typically far from optimal due to
the conservativeness needed to address potentially adversarial instances.
In contrast, machine learning (ML) based optimizers can improve
the average performance by exploiting 
rich historical data and statistical information
\cite{SOCO_OnlineOpt_UnreliablePrediction_Adam_Sigmetrics_2023_10.1145/3579442,SOCO_MetricUntrustedPrediction_Google_ICML_2020_pmlr-v119-antoniadis20a,Shaolei_L2O_ExpertCalibrated_SOCO_SIGMETRICS_Journal_2022,SOCO_ML_ChasingConvexBodiesFunction_Adam_COLT_2022,optimal_christianson2023}, but they sacrifice the strong robustness in terms of provable competitive bounds needed by safety-critical applications, especially when there is  a distributional shift
\cite{L2O_Adversarial_Robust_GAN_arXiv_2020,L2O_AdversarialOnlineResource_ChuanWu_HKU_TOMPECS_2021_10.1145/3494526},
the ML model capacity is limited, and/or inputs are adversarial \cite{L2O_Survey_Amortized_Continuous_Brandon_arXiv_2022_DBLP:journals/corr/abs-2202-00665,L2O_Survey_Benchmark_ZhangyangWang_WotaoYin_arXiv_2021_chen2021learning}.

 More recently,  \emph{ML-augmented online algorithms} have emerged as potential game changers in  classic online problems such as ski rental and caching systems \cite{OnlineOpt_ML_Augmented_RobustCache_Google_ICML_2021_pmlr-v139-chledowski21a,L2O_LearningMLAugmentedAlgorithm_Harvard_ICML_2021_pmlr-v139-du21d,OnlineOpt_Learning_Augmented_RobustnessConsistency_NIPS_2020,OnlineOpt_DataDrivenAlgorithmDesign_Roughgarden_Stanford_CACM_2020_10.1145/3394625,OnlineOpt_ML_Adivce_Survey_2016_10.1145/2993749.2993766}. The goal is to obtain the best of both worlds by utilizing good ML predictions to improve the average performance while ensuring 
bounded competitive ratios even when ML predictions are arbitrarily bad.   
In the context of SOCO, 
there has been initial progress on ML-augmented algorithms 
in the past year \cite{SOCO_MetricUntrustedPrediction_Google_ICML_2020_pmlr-v119-antoniadis20a,SOCO_OnlineOpt_UnreliablePrediction_Adam_Sigmetrics_2023_10.1145/3579442,SOCO_ML_ChasingConvexBodiesFunction_Adam_COLT_2022,Shaolei_SOCO_RobustLearning_OnlineOpt_MemoryCosts_Infocom_2023}.
However, these studies target
the simplest case of SOCO where there is no feedback delay and
the switching costs are linear. Crucially, their specific designs make it  difficult, if not impossible,
to apply to more general and practical settings
where there is hitting cost feedback delay and multi-step nonlinear memory in the switching cost. In addition, with a few exceptions \cite{Shaolei_L2O_ExpertCalibrated_SOCO_SIGMETRICS_Journal_2022,Shaolei_SOCO_RobustLearning_OnlineOpt_MemoryCosts_Infocom_2023},
 a common assumption in the existing SOCO studies is that the  ML model 
is pre-trained as a black box without
awareness of the downstream operation, which creates
a mismatch between training and testing and degrades the average performance.

Even without  ML predictions,
addressing the hitting cost feedback delay and multi-step nonlinear memory is already challenging, as the agent must make decisions  semi-\emph{blindly} without receiving the immediate hitting cost feedback and
the decision at each step can affect multiple future decisions in a complex manner due to multi-step nonlinear switching costs \cite{SOCO_Memory_Adam_NIPS_2020_NEURIPS2020_ed46558a,SOCO_Memory_FeedbackDelay_Nonlinear_Adam_Sigmetrics_2022_10.1145/3508037}. Incorporating ML predictions into the decision process  adds substantial challenges, requiring novel algorithmic techniques beyond those used in the simple SOCO setting with single-step memory and no feedback delay \cite{SOCO_MetricUntrustedPrediction_Google_ICML_2020_pmlr-v119-antoniadis20a,SOCO_OnlineOpt_UnreliablePrediction_Adam_Sigmetrics_2023_10.1145/3579442,SOCO_ML_ChasingConvexBodiesFunction_Adam_COLT_2022}.

\textbf{Contributions.} We propose 
a novel ML-augmented algorithm, called Robustness-Constrained Learning
(\ouralg) that, {for the first time}, provides  both robustness guarantees and
good average performance for SOCO
in general settings with hitting cost feedback delay
and multi-step nonlinear memory in the switching cost.
The foundation of \ouralg is to 
utilize an existing online algorithm (referred to as expert) as well
as a novel {reservation} cost to hedge against future risks while closely following
the ML predictions.  
Without having the immediate hitting cost feedback, \ouralg robustifies untrusted ML predictions at each step
 by judiciously accounting for the 
 hitting cost uncertainties and the non-linear impact of the current decision on future switching costs. 
Importantly, by design, the resulting cost of \ouralg is no greater than $(1+\lambda)$ times the expert's cost for any $\lambda>0$ and any problem instance, while a larger $\lambda$ allows \ouralg to better explore the potential of good ML predictions.

Our main technical results provide bounds on both the worst-case and average-case performance of \ouralg.  In particular, we prove a novel worst-case cost bound on \ouralg in \Cref{thm:setup_feedback_delay} and a bound on the average-case performance in \Cref{thm:average_non_train}.  Our cost bound is proven by utilizing a new
 reservation cost as the core of \ouralg. 
The form of the reservation cost allows us to develop a new proof approach (potentially of independent interest) that decouples the dependency of the online action on the history and constructs a new sufficient robustness constraint to  bound the distance between the actions and ML predictions.  
Importantly, this approach enables us to 
account for multi-step non-linear memory in
the switching cost
and arbitrarily delayed hitting cost feedback in SOCO, which cannot be addressed by the existing algorithms \cite{SOCO_MetricUntrustedPrediction_Google_ICML_2020_pmlr-v119-antoniadis20a,SOCO_OnlineOpt_UnreliablePrediction_Adam_Sigmetrics_2023_10.1145/3579442,SOCO_ML_ChasingConvexBodiesFunction_Adam_COLT_2022,Shaolei_L2O_ExpertCalibrated_SOCO_SIGMETRICS_Journal_2022,Shaolei_SOCO_RobustLearning_OnlineOpt_MemoryCosts_Infocom_2023}. We also provide a first-of-its-kind condition for
simultaneously achieving
both finite robustness and 1-consistency, which has been shown
to be impossible in general \cite{SOCO_OnlineOpt_UnreliablePrediction_Adam_Sigmetrics_2023_10.1145/3579442}.
 Finally, we evaluate the performance of \ouralg using a case study focused on battery management in electric vehicle (EV) charging stations.  Our results highlight the advantage of \ouralg in terms of robustness guarantees compared to pure ML-based methods, as well as the benefit of training a \train ML model.

In summary, our work makes  significant contributions to the growing SOCO literature.  First, we propose a novel ML-augmented algorithm
that provides the first worst-case cost bounds in a general SOCO setting with hitting cost feedback delay and multi-step non-linear switching costs. Our algorithm design and analysis (Theorem~\ref{thm:setup_feedback_delay}) are new and significantly differ
from those used in simple SOCO settings \cite{SOCO_MetricUntrustedPrediction_Google_ICML_2020_pmlr-v119-antoniadis20a,SOCO_OnlineOpt_UnreliablePrediction_Adam_Sigmetrics_2023_10.1145/3579442,SOCO_ML_ChasingConvexBodiesFunction_Adam_COLT_2022,Shaolei_L2O_ExpertCalibrated_SOCO_SIGMETRICS_Journal_2022,Shaolei_SOCO_RobustLearning_OnlineOpt_MemoryCosts_Infocom_2023}. Second, we provide a first sufficient condition
under which finite robustness and 1-consistency are simultaneously
achieved (Corollary~\ref{thm:consitency_robustness}). 
 Finally, we introduce and analyze the first algorithm that allows \train training, 
 highlighting its advantage 
 over the commonly-assumed \nontrain training
 in terms of the average cost.

\section{Related Work}\label{sec:related}
SOCO has been actively studied for over a decade under a wide variety of settings
\cite{SOCO_DynamicRightSizing_Adam_Infocom_2011_LinWiermanAndrewThereska,SOCO_CapacityScalingAdaptiveBalancedGradient_Gatech_MAMA_2021_Sigmetrics_2021_10.1145/3512798.3512808,SOCO_Prediction_LinearSystem_NaLi_Harvard_NIPS_2019_10.5555/3454287.3455620,SOCO_Revisiting_Nanjing_NIPS_2021_zhang2021revisiting,SOCO_Memory_Adam_NIPS_2020_NEURIPS2020_ed46558a,SOCO_ML_ChasingConvexBodiesFunction_Adam_COLT_2022,SOCO_NonConvex_Adam_Sigmetrics_2020_10.1145/3379484,SOCO_OBD_Niangjun_Adam_COLT_2018_DBLP:conf/colt/ChenGW18,SOCO_Prediction_Error_Meta_ZhenhuaLiu_SIGMETRICS_2019_10.1145/3322205.3311087,SOCO_Prediction_RegularizedLookAhead_LeiJiao_XiaojunLin_INFOCOM_2021_9488766,Shaolei_L2O_ExpertCalibrated_SOCO_SIGMETRICS_Journal_2022}.
For example,
designed based on classic algorithmic frameworks, expert online algorithms include
 online gradient descent (OGD) \cite{OGD_zinkevich2003online}, online balanced descent (OBD) \cite{SOCO_OBD_Niangjun_Adam_COLT_2018_DBLP:conf/colt/ChenGW18}, regularized OBD (\robd) \cite{SOCO_OBD_R-OBD_Goel_Adam_NIPS_2019_NEURIPS2019_9f36407e,SOCO_Revisiting_Nanjing_NIPS_2021_zhang2021revisiting}, among many others.
These algorithms are judiciously
designed to have bounded competitive ratios and/or regrets, but they may not perform well on typical instances due to the conservative choices necessary to optimize the worst-case performance. 
 Assuming the knowledge of (possibly imperfect) future inputs, algorithms 
include standard receding horizon control (RHC) \cite{SOCO_Prediction_Error_Meta_ZhenhuaLiu_SIGMETRICS_2019_10.1145/3322205.3311087}
committed horizon control (CHC) \cite{SOCO_Prediction_Error_Niangjun_Sigmetrics_2016_10.1145/2964791.2901464},
and receding horizon gradient descent (RHGD) \cite{Receding_Horizon_GD_li2020online,SOCO_Prediction_LinearSystem_NaLi_Harvard_NIPS_2019_10.5555/3454287.3455620}. Nonetheless, the worst-case performance is still unbounded
when the inputs have large errors. 

By tapping into  historical data, 
pure ML-based online optimizers, e.g.,
recurrent neural networks,
have been studied for online problems  
\cite{L2O_NewDog_OldTrick_Google_ICLR_2019,L2O_OnlineBipartiteMatching_Toronto_ArXiv_2021_DBLP:journals/corr/abs-2109-10380,L2O_OnlineResource_PriceCloud_ChuanWu_AAAI_2019_10.1609/aaai.v33i01.33017570}.
Nonetheless, even with (heuristic) techniques such
as distributionally robust training and/or addition
of hard training instances (e.g., adversarial samples) \cite{L2O_AdversarialOnlineResource_ChuanWu_HKU_TOMPECS_2021_10.1145/3494526,L2O_Adversarial_Robust_GAN_arXiv_2020}, 
they cannot provide formal worst-case guarantees as their expert counterparts.

By combining potentially untrusted ML predictions with
robust experts,
ML-augmented algorithms
have emerged as a promising approach
\cite{OnlineOpt_Learning_Augmented_RobustnessConsistency_NIPS_2020,OnlineOpt_ML_Adivce_Survey_2016_10.1145/2993749.2993766,OnlineOpt_ML_Advice_CompetitiveCache_Google_JACM_2021_10.1145/3447579,OnlineOpt_ML_Augmented_RobustCache_Google_ICML_2021_pmlr-v139-chledowski21a}.
The existing ML-augmented algorithms for SOCO 
\cite{Shaolei_SOCO_RobustLearning_OnlineOpt_MemoryCosts_Infocom_2023,SOCO_MetricUntrustedPrediction_Google_ICML_2020_pmlr-v119-antoniadis20a,SOCO_OnlineOpt_UnreliablePrediction_Adam_Sigmetrics_2023_10.1145/3579442,SOCO_ML_ChasingConvexBodiesFunction_Adam_COLT_2022,optimal_christianson2023,Shaolei_L2O_ExpertCalibrated_SOCO_SIGMETRICS_Journal_2022} 
only focus on simple SOCO settings where the hitting cost is known
without delays and the switching cost is linear.
 Extending these algorithms \cite{Shaolei_SOCO_RobustLearning_OnlineOpt_MemoryCosts_Infocom_2023,SOCO_MetricUntrustedPrediction_Google_ICML_2020_pmlr-v119-antoniadis20a,SOCO_OnlineOpt_UnreliablePrediction_Adam_Sigmetrics_2023_10.1145/3579442,SOCO_ML_ChasingConvexBodiesFunction_Adam_COLT_2022,optimal_christianson2023,Shaolei_L2O_ExpertCalibrated_SOCO_SIGMETRICS_Journal_2022} 
to the general SOCO setting requires substantially
new designs and analysis. For example,
\cite{Shaolei_SOCO_RobustLearning_OnlineOpt_MemoryCosts_Infocom_2023}
utilizes the simple triangle inequality for
linear switching costs in the metric
space to achieve robustness, whereas
this inequality does not
hold given (multi-step) non-linear memory
in terms of squared
switching costs \cite{SOCO_OnlineOpt_UnreliablePrediction_Adam_Sigmetrics_2023_10.1145/3579442} even when there is no
feedback delay.
In fact, even without considering ML predictions, the general SOCO setting
with feedback delays and multi-step non-linear switching
costs presents significant challenges that need
new techniques \cite{SOCO_Memory_FeedbackDelay_Nonlinear_Adam_Sigmetrics_2022_10.1145/3508037,SOCO_Memory_Adam_NIPS_2020_NEURIPS2020_ed46558a} beyond those for the simple SOCO setting. Thus, \ouralg makes novel contributions
to the growing ML-augmented SOCO literature.
Customizing ML to better suit the downstream operation to achieve a lower cost has been considered for a few online problems 
 \cite{L2O_LearningMLAugmented_Regression_CR_GeRong_NIPS_2021_anand2021a,Shaolei_L2O_ExpertCalibrated_SOCO_SIGMETRICS_2022,Shaolei_SOCO_RobustLearning_OnlineOpt_MemoryCosts_Infocom_2023}. In \ouralg, however, we need implicit differentiation through time to customize
ML by considering our novel algorithm designs in our general SOCO setting.

In online learning with expert predictions \cite{OnlineOpt_Expert_Bufering_WeightedMajority_geulen2010regret,OnlineOpt_Expert_EasyData_NIPS2014_01f78be6,OnlineOpt_Expert_MemoryBounds_STOC_2022_10.1145/3519935.3520069}, experts are dynamically chosen
 with time-varying probabilities to achieve a low regret compared
to the best expert in hindsight. 
By contrast, \ouralg considers a different problem setting
with feedback
delay and multi-step non-linear memory,
and focuses on constrained
learning by bounding the total cost below
$(1+\lambda)$ times of the expert's cost for any instance
and any $\lambda>0$.
 Finally, \ouralg is also broadly relevant to 
conservative  bandits and reinforcement learning \cite{Conservative_RL_Bandits_LiweiWang_SimonDu_ICLR_2022_yang2022a}. 
Specifically, conservative exploration focuses on unknown
cost functions (and, when applicable, transition models) and
uses a baseline policy to guide the exploration process.
But, its design is fundamentally different in that 
it does not hedge against future uncertainties when choosing an action for each step. 
Additionally, constrained policy optimization
\cite{Conservative_ProjectBasedConstrainedPolicyOptimizatino_ICLR_2020_Yang2020Projection-Based,Conservative_RL_OfflineDistributional_JasonMa_NIPS_2021_ma2021conservative} focuses on constraining the \emph{average} cost, whereas
\ouralg focuses on the worst-case cost constraint.

\section{Model and Preliminaries}
In a SOCO problem, an agent, a.k.a., decision maker, must select an irrevocable action $x_t$ from an action space $\mathcal{X}\subseteq\mathbb{R}^n$ with size $|\mathcal{X}|$ at each of time $t=1,\ldots,T$.  Given the selected action, the agent incurs the sum of (i) a non-negative hitting cost $f(x_t,y_t)\geq0$ parameterized by the context $y_t\in\mathcal{Y}\subseteq\mathbb{R}^m$, where $f(\cdot): \mathbb{R}^n \rightarrow \mathbb{R}_{\geq 0}$, and (ii) a non-negative switching cost $d(x_t,x_{t-p:t-1})=\frac{1}{2}\|x_t-\delta(x_{t-p:t-1})\|^2$,
where the constant $\frac{1}{2}$ is added
for the convenience of derivation, $\|\cdot\|$ is the $l_2$ norm by default,
and $\delta(\cdot):\mathbb{R}^{p\times n} \rightarrow\mathbb{R}^{n}$ is a (possibly non-linear) function of $x_{t-p:t-1}=(x_{t-p},\cdots,x_{t-1})$. We make the following standard assumptions.

\begin{assumption}\label{assumption:hitting}
At each $t$, the hitting cost 
$f(x_t,y_t)$ is non-negative, $\alpha_h$-strongly convex, and $\beta_h$-smooth
in $x_t\in\mathcal{X}$. It is also Lipschitz continuous with respect to $y_t\in\mathcal{Y}$.
\end{assumption}

\begin{assumption}\label{assumption:switching}
In the switching cost 
$d(x_t,x_{t-p:t-1})=\frac{1}{2}\|x_t-\delta(x_{t-p:t-1})\|^2$, the function
 $\delta(x_{t-p:t-1})$ is $L_i$-Lipschitz continuous in $x_{t-i}$ for $i=1\cdots p$, i.e., for any
 $x_{t-i}, x_{t-i}' \in \mathcal{X}$, we have
 $\| \delta(x_{t-p}, \cdots, x_{t-i}, \cdots x_{t-1}) - \delta(x_{t-p}, \cdots, x_{t-i}', \cdots x_{t-1}) \| 
        \leq  L_i \| x_{t-i} - x_{t-i}' \|$.
\end{assumption}

The convexity of the hitting cost is standard in the literature
and needed for competitive
analysis,
while smoothness (i.e., 
Lipschitz-continuous gradients)
guarantees that bounded action differences also result
in bounded cost differences \cite{SOCO_Memory_FeedbackDelay_Nonlinear_Adam_Sigmetrics_2022_10.1145/3508037}. A common example
of the hitting cost is $f(x_t,y_t)=\|x_t-y_t\|^2$
as motivated by object tracking applications, where
$y_t$ is the online moving target \cite{SOCO_Memory_Adam_NIPS_2020_NEURIPS2020_ed46558a,SOCO_Memory_FeedbackDelay_Nonlinear_Adam_Sigmetrics_2022_10.1145/3508037}.
 In the  
switching cost term, 
the previous $p$-step actions $x_{t-p:t-1}$ are essentially
encoded by 
  $\delta(x_{t-p:t-1})$
 \cite{SOCO_Memory_Adam_NIPS_2020_NEURIPS2020_ed46558a}.
Let us take drone tracking as an example. The switching cost
can be written as $d(x_t,x_{t-1})=\frac{1}{2}\|x_t-x_{t-1}+C_1+C_2\cdot|x_{t-1}|\cdot x_{t-1}\|^2$ and hence $\delta(x_{t-1})=x_{t-1}-C_1-C_2\cdot|x_{t-1}|\cdot x_{t-1}$ is nonlinear,
where $x_t$ is the drone's speed at time $t$
and the constants of $C_1$ and $C_2$ account for gravity and the aerodynamic drag
\cite{ML_DroneLander_GuanyaShi_Yisong_ICRA_2019_shi2019neural}.
For additional examples of switching costs in other applications,
readers are further referred
to \cite{SOCO_Memory_FeedbackDelay_Nonlinear_Adam_Sigmetrics_2022_10.1145/3508037}.

For the convenience of presentation, we 
use ${y}=(y_1,\cdots,y_T)\in\mathcal{Y}^T$ to denote a problem instance, while noting that the initial actions $x_{-p+1:0}=(x_{-p+1},\cdots,x_0)$ are also provided
as an additional input.  Further, 
for
$1\leq t_1\leq t_2\leq T$,
we also rewrite $\sum_{\tau=t_1}^{t_2}f(x_{\tau},y_{\tau})+d(x_{\tau},x_{{\tau}-p:{\tau}-1})$ as $\mathrm{cost}(x_{t_1:t_2})$,
where we suppress the context  $y_t$ without ambiguity.

For online optimization, the key challenge is that the switching cost couples online actions
and the hitting costs are revealed online.
As in the recent literature on SOCO
 \cite{SOCO_Memory_FeedbackDelay_Nonlinear_Adam_Sigmetrics_2022_10.1145/3508037},
we assume that
the agent knows the switching cost, because it is determined
endogenously by the problem itself and the agent's previous actions.
The agent also knows the smoothness constant $\beta_h$,
although the hitting cost function itself is revealed online
subject to a delay as defined below.

\subsection{Feedback Delay}\label{sec:formulation_feedback_delay}
There may be feedback delay that prevents immediate observation of the context
$y_t$ (which is equivalent to delayed
hitting cost function)  \cite{Control_CompetitiveControl_Memory_InexactPrediction_GuanyaShi_CISS_2021_shi2021competitive,SOCO_Memory_Adam_NIPS_2020_NEURIPS2020_ed46558a}. 
For example, in assisted drone landing,
the context parameter $y_t$ can represent the target velocity
at time $t$ sent to the drone by a control center, but the communications
between the drone and control center can experience delays
due to wireless channels and/or even packet losses due to adversarial jamming  \cite{SOCO_Memory_FeedbackDelay_Nonlinear_Adam_Sigmetrics_2022_10.1145/3508037,ML_DroneLander_GuanyaShi_Yisong_ICRA_2019_shi2019neural}.

To model the delay,
we refer to $q\geq0$ as the maximum feedback delay
(i.e., context $y_t$ can be delayed for up to $q\geq0$ steps), and 
define $q$-delayed time set of arrival contexts.

\begin{definition}[$q$-delayed time set
of arrival contexts]\label{definition:time_set}
Given the maximum feedback delay of  $q\geq0$, 
for each time $t=1,\ldots, T$, the $q$-delayed time set
of arrival contexts contains the time
 indexes whose contexts are newly revealed to the
agent at time $t$
and is defined as
$\mathcal{D}_t^q\subseteq\{\tau\in\mathbb{N}\;|t-q\leq\tau\leq t\}$ such that
$\{\tau\in\mathbb{N}\;|\tau\leq t-q\}\subseteq \left(\mathcal{D}_1^q\bigcup\cdots\bigcup \mathcal{D}_{t}^q\right)$.
\end{definition}

Naturally, given the maximum delay
of $q\geq0$, we must have $\{\tau\in\mathbb{N}\;|
 \tau\leq t-q\}\subseteq \left(\mathcal{D}_1^q\bigcup\cdots\bigcup \mathcal{D}_{t}^q\right)$, i.e., at time $t$, the agent must have
already known the contexts $y_{\tau}$ for any $\tau=1,\cdots, t-q$. 

It is worth highlighting that our definition of
$\mathcal{D}_t^q$ is flexible and applies to various
delay models. Specifically, the no-delay setting corresponds to $q=0$ and $D_t^{q=0}=\{t\}$, while $q=T$ captures the case in which the agent may potentially have to choose actions without knowing
any of the contexts $y_1,\cdots,y_T$ throughout an entire problem instance.
Given the maximum delay $q\geq0$, the delayed contexts can be revealed to the agent in various orders different from  their actual time steps, 
i.e., the agent may receive $y_{t_1}$ earlier than $y_{t_2}$ for $t_1>t_2$. 
Also, the agent can receive a batch of contexts $y_{t-q},\cdots,y_t$ simultaneously at time $t$, and receive no new contexts some other time steps. 

In online optimization, handling delayed cost functions, even for a single step, is challenging  \cite{SOCO_Memory_Adam_NIPS_2020_NEURIPS2020_ed46558a,ML_OnlineLearn_LocalPermutation_Delay_ICML_2017_10.5555/3305890.3306000,SOCO_Memory_FeedbackDelay_Nonlinear_Adam_Sigmetrics_2022_10.1145/3508037}.
Adding ML predictions into online optimization
creates further  algorithmic difficulties.

\subsection{Performance Metrics}
Our goal is to minimize the sum of the total hitting costs and switching costs over $T$ time steps: $\min_{x_1,\cdots x_T} \sum_{t=1}^Tf(x_t,y_t)+d(x_t,x_{t-p:t-1})$. We formally define our two measures of interest.

\begin{definition}[Competitiveness]\label{definition:cr}
An algorithm $ALG_1$ is said to be $CR$-competitive against
another baseline algorithm $ALG_2$ if 
${cost}(ALG_1,{y})\leq CR\cdot {cost}(ALG_2,{y})$ is satisfied
for any problem instance $y\in\mathcal{Y}^T$,
where  ${cost}(ALG_1,{y})$ and $\mathrm{cost}(ALG_2,{y})$ denote the total costs of $ALG_1$ and $ALG_2$, respectively.
\end{definition}

\begin{definition}[Average cost]\label{definition:average}
The average cost
 of an  algorithm $ALG$ is 
$\overline{{cost}}(ALG)=\mathbb{E}_{{y}\in \mathbb{P}_{{y}}}\left[
{cost}(ALG,{y})\right]$, 
where ${cost}(ALG,{y})$ denotes
the cost of $ALG$ for a problem instance ${y}$,
and $\mathbb{P}_{y}$ is the  exogenous probability %density
distribution
of  ${y}=(y_1,\cdots,y_T)\in\mathcal{Y}^T$.
\end{definition}

Our definition of competitiveness against
a general baseline algorithm is commonly considered in the literature, e.g., \cite{SOCO_ML_ChasingConvexBodiesFunction_Adam_COLT_2022}.
 The two metrics measure an online algorithm's
 robustness in the worst case
 and expected performance in the average case,
 which are both important in practice.

We consider an
expert (online) algorithm $\pi$ which chooses $x_t^{\pi}$ at time $t$
and an ML model $h_W$ which, parameterized by $W$,
produces $\tilde{x}_t=h_W\left(\tilde{x}_{t-p:t-1}, \left\{y_{\tau}|\tau\in\mathcal{D}_t^q\right\}\right)$ at time $t$. 
 As in the existing ML-augmented online algorithms \cite{SOCO_ML_ChasingConvexBodiesFunction_Adam_COLT_2022,SOCO_OnlineOpt_UnreliablePrediction_Adam_Sigmetrics_2023_10.1145/3579442},
\ouralg chooses an  actual online action 
 $x_t$ by using the two actions $x_t^{\pi}$
and $\tilde{x}_t$ as advice. In general, it is extremely challenging, if not impossible, to simultaneously optimize for both the average cost and the competitiveness. 
Here, given a robustness requirement $\lambda>0$,
we focus on minimizing the average cost while
ensuring $(1+\lambda)$-competitiveness against the expert
$\pi$. Crucially, the optimal expert for our setting is iROBD,
which has the best-known competitiveness against the offline optimal
$OPT$ with complete information
\cite{SOCO_Memory_FeedbackDelay_Nonlinear_Adam_Sigmetrics_2022_10.1145/3508037}. Thus, by using iROBD as the expert $\pi$, 
 $(1+\lambda)$-competitiveness of \ouralg
against $\pi$ 
 immediately translates into a scaled competitive ratio
 of $(1+\lambda)\cdot CR_{iROBD}$
 against $OPT$, where $CR_{iROBD}$ is iROBD's competitive ratio against $OPT$.

\section{\ouralg: The Design and Analysis}\label{sec:single_switching}

In this section, we 
present \ouralg, 
 a novel ML-augmented
 that combines ML predictions (i.e.,
 online actions produced by an ML model \cite{SOCO_MetricUntrustedPrediction_Google_ICML_2020_pmlr-v119-antoniadis20a,SOCO_OnlineOpt_UntrustedPredictions_Switching_Adam_arXiv_2022,Shaolei_L2O_ExpertCalibrated_SOCO_SIGMETRICS_Journal_2022, optimal_christianson2023}) with a robust expert online algorithm 
 to 
the worst-case cost while leveraging the benefit of ML predictions for average performance.

\subsection{Robustness-Constrained Online Algorithm}\label{sec:single_algorithm}

Our goal is to ``robustify'' ML predictions, by which we mean that we want to ensure a robustness bound on the cost of no greater than 
$(1+\lambda)$ times of the expert's  cost,
i.e., for any problem instance $y$, we have $\mathrm{cost}(x_{1:T})\leq (1+\lambda)\mathrm{cost}(x_{1:T}^{\pi})$, where $\lambda>0$ is
a hyperparameter indicating the level of robustness we would like to achieve. Meanwhile, we would like to utilize the benefits of ML predictions to improve
the average performance.

Because of the potential feedback delays, \ouralg needs to choose
an online action $x_t$ without necessarily knowing the hitting costs of the expert's action $x_t^{\pi}$ and ML prediction $\tilde{x}_t$.
Additionally, the action $x_t$ can affect multiple future switching costs
due to multi-step non-linear memory.
Thus, it is very challenging
to robustify ML predictions for the SOCO settings we consider. 
 A simple approach 
one might consider is to constrain
${x}_t$ such that the cumulative cost up to each time $t$
is always no greater than $(1+\lambda)$ times of the expert's cumulative cost,
i.e., $\mathrm{cost}(x_{1:t})\leq (1+\lambda)\mathrm{cost}(x_{1:t}^{\pi})$.
However, even without feedback delays, such an approach may not  even produce feasible actions for some $t=1,\cdots,T$.
 We explain this by considering a single-step switching cost case.
Suppose that $\mathrm{cost}(x_{1:t})\leq (1+\lambda)\mathrm{cost}(x_{1:t}^{\pi})$
is satisfied at a time $t< T$, and we choose an action $x_t\not=x_t^{\pi}$
different from the expert. Then, at
time $t+1$, let us consider a case in which the expert algorithm has such a low cost that even choosing $x_{t+1}=x_{t+1}^{\pi}$
will result in $\mathrm{cost}(x_{1:t})+f(x_{t+1}^{\pi},y_{t+1})+d(x_{t+1}^{\pi},
x_t)> (1+\lambda)\left[\mathrm{cost}(x_{1:t}^{\pi})+f(x_{t+1}^{\pi},y_{t+1})+d(x_{t+1}^{\pi},
x_t^{\pi})\right]$. This is because the actual switching cost
$d(x_{t+1}^{\pi},
x_t)$ can be significantly higher than the expert's switching
cost $d(x_{t+1}^{\pi},
x_t^{\pi})$ due to $x_t\not=x_t^{\pi}$. 
As a result, at time $t+1$, it is possible that there exist
no feasible actions that satisfy 
 $\mathrm{cost}(x_{1:t+1})\leq (1+\lambda)\mathrm{cost}(x_{1:t+1}^{\pi})$.
Moreover, when choosing an online action
close to the ML prediction, extra caution must be exercised as the hitting
costs can be revealed with delays of up to $q$ steps.

To address these challenges, \ouralg introduces  novel reservation
costs to hedge against any possible uncertainties due to hitting cost feedback delays
and multi-step non-linear memory in the switching costs. Concretely,
given both the expert action
$x_t^{\pi}$ and ML prediction
$\tilde{x}_t$ at time $t$, 
we choose $x_t  = \arg\min_{x\in\mathcal{X}_t} \frac{1}{2} \| x - \tilde{x}_t \| ^2$
by solving
a constrained convex problem to
project the ML prediction $\tilde{x}_t$ into a robustified action set
$x\in\mathcal{X}_t$ where $x_t$ satisfies:
\begin{equation}\label{eqn:proj_multi_delay_constraint}
\begin{split}
    & \sum_{\tau\in\mathcal{A}_t}f(x_{\tau}, y_\tau)+\sum_{\tau=1}^td(x_{\tau}, x_{\tau-p:\tau-1}) +\sum_{\tau\in\mathcal{B}_t}H(x_\tau, x_\tau^\pi)+ G(x_t, x_{t-p:t-1}, x_{t-p:t}^\pi)\\
    \leq&(1+\lambda)\left( \sum_{\tau\in\mathcal{A}_t}f(x^{\pi}_{\tau}, y_\tau)
  + \sum_{\tau=1}^td(x_{\tau}^\pi, x_{\tau-p:\tau-1}^\pi) \right),
    \end{split}
\end{equation}
in which $\lambda> 0$
is the robustness hyperparameter,
$\mathcal{A}_t^q=\left(\mathcal{D}_1^q\bigcup\cdots\bigcup \mathcal{D}_{t}^q\right)$ and $\mathcal{B}_t^q=\{1,\cdots,t\}\backslash\mathcal{A}_t$
are the sets of time  indexes for which the agent knows and does not know the context parameters up to time $t$, respectively.
Most importantly, the two novel reservation costs
$H(x_\tau, x_\tau^\pi)$ and 
$G(x_t, x_{t-q:t-1}, x_{t-q:t}^\pi)$ are defined
as 
\begin{equation}\label{eqn:reservation_hitting_delay}
     H(x_{\tau}, x_{\tau}^\pi) = \frac{\beta_h}{2}(1+ \frac{1}{\lambda_0}) \|x_{\tau} - x_{{\tau}}^\pi \|^2,
\end{equation}
\vspace{-0.5cm}
\begin{equation}\label{eqn:reservation_multi}
\begin{split}
    &G(x_t, x_{t-p:t-1}, x_{t-p:t}^\pi)
    = 
\frac{(1 + \frac{1}{\lambda_0})\left(1+ \sum_{k=1}^{p}L_k\right)}{2}\sum_{k=1}^{p}\biggl(L_k \| x_t - x_t^\pi\|^2 + \sum_{i=1}^{p-k}L_{k+i} \| x_{t-i} - x_{t-i}^\pi\|^2 \biggr),
\end{split}
\end{equation}
where  $\beta_h$ is the smoothness constant of the hitting cost 
in Assumption~\ref{assumption:hitting},
$L_k$
is the Lipschitz constant of $\delta(x_{t-p:t-1})$
in Assumption~\eqref{assumption:switching}, and 
$\lambda_0 \in (0, \lambda)$ with the optimum
being $\lambda_0=\sqrt{1+\lambda}-1$ (Theorem~\ref{thm:setup_feedback_delay}). The computational complexity
for projection into  \eqref{eqn:proj_multi_delay_constraint} is tolerable
due to convexity. 

The interpretation of $H(x_\tau, x_\tau^\pi)$ and 
$G(x, x_{t-q:t-1}, x_{t-q:t}^\pi)$ is as follows.
If \ouralg's action $x_{\tau}$ deviates from the expert's action $x_{\tau}^{\pi}$
at time ${\tau}$ and the hitting cost is not known yet due to delayed $y_{\tau}$,
then it is possible that \ouralg actually experiences
a high but unknown hitting cost.
In this case, to guarantee the worst-case robustness, we include an upper bound
of the cost difference as the reservation cost
such that $f(x_{\tau},y_{\tau})- (1+\lambda)f(x_{\tau}^{\pi},y_{\tau})
 \leq H(x_{\tau}, x_{\tau}^\pi)$
 regardless of the delayed $y_{\tau}\in\mathcal{Y}$.
 If $y_\tau$ has been revealed at time $t$ (i.e.,
 $\tau\in\mathcal{A}_t$), then we use the actual costs
 instead of the reservation cost.
Likewise, by considering the expert's future actions
 as a feasible backup plan in the worst case, the reservation cost 
 $G(x, x_{t-p:t-1}, x_{t-p:t}^\pi)$ upper bounds
 the maximum possible difference in the future switching costs (up
 to future $p$ steps)
 due to deviating from the expert's action at time $t$.
 
With $H(x_\tau, x_\tau^\pi)$ and 
$G(x, x_{t-q:t-1}, x_{t-q:t}^\pi)$ as reservation costs in \eqref{eqn:proj_multi_delay_constraint}, \ouralg achieves
robustness by ensuring that following the expert's actions in the future is always feasible, regardless of the delayed $y_t$. 
 The online algorithm is  described
in Algorithm~\ref{alg:RP-OBD_delay}, where
both the expert and ML model
have the same online information $\left\{y_{\tau}|\tau\in\mathcal{D}_t^q\right\}$ at time $t$
and produce their own actions as advice to \ouralg.

\begin{algorithm}[!t]
\caption{Online Optimization with \ouralg}
\begin{algorithmic}[1]\label{alg:RP-OBD_delay}
\REQUIRE $\lambda>0$, $\lambda_0\in(0, \lambda)$,  initial actions $x_{1-p:0}$, expert algorithm $\pi$, and ML model $h_W$
\STATE for $t=1,\cdots, T$
\STATE \quad Receive a set of contexts $\left\{y_{\tau}| \tau\in\mathcal{D}_t^q\right\}$
\STATE \quad Get the expert's action $x_t^\pi$ given its own history
\STATE \quad Get $\tilde{x}_t = h_W\left(\tilde{x}_{t-p:t-1}, \left\{y_{\tau}|\tau\in\mathcal{D}_t^q\right\}\right)$  %given its own history%\texttt{//ML prediction}
\STATE \quad Choose  $x_t  = \arg\min_{x\in\mathcal{X}_t} \frac{1}{2} \| x - \tilde{x}_t \| ^2$ subject to the constraint~\eqref{eqn:proj_multi_delay_constraint}
\;\;\texttt{//Robustification}
\end{algorithmic}
\end{algorithm}

\subsection{Analysis}\label{sec:single_cost_ratio}

We now present our main results on the cost bound of \ouralg,
showing that \ouralg can indeed maintain the desired $(1+\lambda)$-competitiveness against the expert $\pi$ while exploiting the potential of ML predictions.

\begin{theorem}[Cost bound] \label{thm:setup_feedback_delay}
Consider a memory length $p\geq1$
and the maximum feedback delay of $q\geq0$.
Given a context sequence $y=(y_1,\cdots,y_T)$, let
$\text{cost}(\tilde{x}_{1:T})$ and $\text{cost}({x}^{\pi}_{1:T})$ be
the costs of pure ML predictions $\tilde{x}_{1:T}$
and expert actions ${x}_{1:T}^\pi$, respectively.
 For any $\lambda>0$,
by optimally setting $\lambda_0 = \sqrt{1+\lambda} - 1$, the cost of \ouralg
is upper bounded by
\begin{equation}\label{eqn:cr_bound_delay}
  \begin{split}
  \text{cost}(x_{1:T}) \leq \min \Biggl((1+\lambda) \text{cost}({x}_{1:T}^\pi), \biggl(\sqrt{\text{cost}(\tilde{x}_{1:T})} + \sqrt{\frac{\beta_h+ \alpha^2}{2} \Delta(\lambda)} \biggr)^2 \Biggr),
   \end{split}
\end{equation}
where  $\Delta(\lambda) = \sum_{i=1}^T   \Bigl[   \| \tilde{x}_t  - x_t^\pi\|^2 -  \frac{2(\sqrt{1+\lambda}-1)^2}{(\beta_h+ \alpha^2) }   \text{cost}_t^\pi   \Bigr]^+$ in which
$\text{cost}_t^\pi=\left(\sum_{\tau\in\mathcal{D}_t^q}f(x_{\tau}^\pi, y_{\tau})\right) + d(x_t^{\pi}, x_{t-p:t-1}^{\pi})$ is
the total of revealed hitting costs and switching cost
for the expert at time $t$,
$\beta_h$ is the smoothness constant of the hitting cost
(Assumption~\ref{assumption:hitting}), and
 $\alpha=1+\sum_{i=1}^pL_i$ with
$L_1\dots L_p$ being the Lipschitz constants in
the switching cost (Assumption~\ref{assumption:switching}).
\hfill$\square$
\end{theorem}

Theorem~\ref{thm:setup_feedback_delay} is the \emph{first} worst-case cost
analysis for ML-augmented algorithms in a general SOCO setting
with delayed hitting costs and multi-step switching costs.
Its proof is available
in the appendix and outlined here. We prove the competitiveness against
the expert $\pi$ based on
our novel reservation cost $H(x_\tau, x_\tau^\pi)$ and 
$G(x_t, x_{t-q:t-1}, x_{t-q:t}^\pi)$  by induction.
It is more challenging to prove the competitiveness against the ML prediction,
because $x_t$ implicitly depends on all the previous actions of ML predictions and expert actions up to time $t$.
To address this challenge, we utilize a novel technique by first removing the dependency of $x_t$ on the history. 
Then, we construct a new sufficient robustness constraint 
that allows an explicit expression of another robustified
action, whose distance to the ML prediction $\tilde{x}_t$
is an upper bound by the distance between $x_t$ and $\tilde{x}_t$
due to the projection of $\tilde{x}_t$ into \eqref{eqn:proj_multi_delay_constraint}.
Finally, due to the smoothness of the hitting cost function and the switching cost,
the distance bound 
translates into the competitiveness of \ouralg against
ML predictions.

 The two terms inside the $\min$ operator in Theorem~\ref{thm:setup_feedback_delay} show
 the tradeoff between achieving better competitiveness
 and more closely following ML predictions.
To interpret this, we note  that the first term inside $\min$ operator 
shows  $(1+\lambda)$-competitiveness of \ouralg against the expert
$\pi$,  while the second term inside $\min$ operator shows 
\ouralg can also exploit the potential of good ML predictions.
A smaller $\lambda>0$
means that we want to be closer to the expert for better competitiveness,
while
 a larger $\lambda>0$  decreases
 the term $\Delta(\lambda) = \sum_{i=1}^T   \Bigl[   \| \tilde{x}_t  - x_t^\pi\|^2 -  \frac{2(\sqrt{1+\lambda}-1)^2}{(\beta_h+ \alpha^2) }   \text{cost}_t^\pi   \Bigr]^+$
 and hence
makes \ouralg follow the ML predictions more closely.

The term $\Delta(\lambda)$ in \eqref{eqn:cr_bound_delay} essentially
bounds the total squared distance between the actual
online action and ML predictions.
Intuitively, 
 \ouralg should follow 
the ML predictions more aggressively when the expert
does not perform well.
 This insight is also
reflected in $\Delta(\lambda)$ in Theorem~\ref{thm:setup_feedback_delay}. Concretely,
when
the expert $\pi$'s total revealed 
${cost}_t^\pi$ is higher, $\Delta(\lambda)$ also becomes smaller,
pushing \ouralg closer to ML.
On the other hand, when the expert's cost is lower,
\ouralg stays closer to the better-performing expert for guaranteed competitiveness.
 When $\| \tilde{x}_t  - x_t^\pi\|^2$ is
larger (i.e., greater discrepancy
between the expert's action and ML prediction),
it is naturally more difficult to follow
both the expert and ML prediction simultaneously.
Thus, given a robustness requirement of $\lambda>0$,
we see from $\Delta(\lambda)$ that a larger
$\| \tilde{x}_t  - x_t^\pi\|^2$ also increases the second
term in the $\min$ operator in Theorem~\ref{thm:setup_feedback_delay},
making it more difficult for \ouralg to exploit the potential
of ML predictions. Moreover, deviating from the
expert's action at one step can have impacts on
the switching costs in future $p$
steps. 
Thus, the  memory length $p$  creates some additional friction
 for \ouralg to achieve a low cost bound with respect to ML predictions:
 the greater $p$, the greater $\alpha=1+\sum_{i=1}^pL_i$, and hence the greater the second term in
\eqref{eqn:cr_bound_delay}.

\textbf{Robustness and consistency.}
It remains to show the worst-case competitiveness against the
offline optimal algorithm $OPT$,
which is typically performed 
for two extreme cases --- when
 ML predictions are extremely bad 
and perfect --- which are respectively referred
to as \emph{robustness} and \emph{consistency} 
in the literature  \cite{SOCO_MetricUntrustedPrediction_Google_ICML_2020_pmlr-v119-antoniadis20a,OnlineOpt_ML_Augmented_RobustCache_Google_ICML_2021_pmlr-v139-chledowski21a,OnlineOpt_Learning_Augmented_RobustnessConsistency_NIPS_2020} and formally defined below.

\begin{definition}[Robustness and consistency]\label{definition:robustness_consistency}
The robustness of \ouralg
is $CR(\infty)$ if \ouralg is $CR(\infty)$-competitive
against $OPT$ when the ML's competitiveness against
$OPT$ is arbitrarily large (denoted
as $\tilde{CR}\to\infty$) ;
and the consistency of \ouralg
is $CR(1)$ if \ouralg is $CR(1)$-competitive
against $OPT$ 
when the ML's competitiveness against
$OPT$ is $\tilde{CR}=1$.
\end{definition}

Robustness indicates the worst-case performance of \ouralg
for any possible ML predictions, whereas consistency measures the capability of \ouralg
to retain the performance of perfect ML predictions.
In general, the tradeoff between robustness and consistency is  unavoidable for online algorithms
\cite{OnlineOpt_ML_Augmented_RobustCache_Google_ICML_2021_pmlr-v139-chledowski21a,SOCO_OnlineOpt_UnreliablePrediction_Adam_Sigmetrics_2023_10.1145/3579442,Control_RobustConsistency_LQC_TongxinLi_Sigmetrics_2022_10.1145/3508038}. The state-of-the-art expert algorithm iROBD recently proposed
in \cite{SOCO_Memory_FeedbackDelay_Nonlinear_Adam_Sigmetrics_2022_10.1145/3508037}
has the best-known competitive ratio under
the assumption of identical delays for each context
(i.e., $\mathcal{D}_t^q=\{t-q\}$ --- each context is delayed
by $q$ steps).
 The identical-delay model
essentially ignores
any other contexts $y_{\tau}$ for
 $\tau\in\{\tau\in\mathbb{N}\;|t-q+1\leq\tau\leq t\}$.
Thus, it is the worst case of a general $q$-step delay
setting, whose competitive ratio is upper bounded
by that of iROBD.
Consequently, 
due to $(1+\lambda)$-competitiveness against
any expert $\pi$ in Theorem~\ref{thm:setup_feedback_delay},
we immediately obtain a finite robustness bound for \ouralg by considering
iROBD as the expert. 

Nonetheless,
even for the simplest SOCO setting
with no feedback delay and a switching cost of $d(x_t,x_{t-1})=\frac{1}{2}\|x_t-x_{t-1}\|^2$,
a recent study  \cite{SOCO_OnlineOpt_UnreliablePrediction_Adam_Sigmetrics_2023_10.1145/3579442} has shown that
it is \emph{impossible} to simultaneously achieve 1-consistency and
finite robustness.
Consequently, in general SOCO settings,
the finite robustness of \ouralg given
any $\lambda>0$ means the impossibility of achieving 1-consistency by following perfect ML predictions without further assumptions. 

Despite this pessimistic result due to
the fundamental challenge of SOCO, 
we find a sufficient condition that can overcome
the impossibility, which is formalized as follows.

\begin{corollary}[1-consistency and finite robustness] \label{thm:consitency_robustness}
Consider iROBD 
as the exert $\pi$, whose competitive
ratio against $OPT$ is denoted as
$CR_{iROBD}$ \cite{SOCO_Memory_FeedbackDelay_Nonlinear_Adam_Sigmetrics_2022_10.1145/3508037}.
If the expert's switching cost always satisfies
$d(x_{t}^\pi, x_{t-p:t-1}^\pi)\geq\epsilon>0$
for any time $t=1,\cdots,T$,
then by setting $\lambda\geq \frac{|\mathcal{X}|^2(\alpha^2+\beta_h)}{2\epsilon}+\sqrt{\frac{2|\mathcal{X}|^2(\alpha^2+\beta_h)}{\epsilon}}\sim\mathcal{O}(\frac{1}{\epsilon})$
and optimally using $\lambda_0 = \sqrt{1+\lambda} - 1$,
\ouralg achieves $(1+\lambda)\cdot CR_{iROBD}$-robustness
and 1-consistency simultaneously.
\hfill$\square$
\end{corollary}

Corollary~\ref{thm:consitency_robustness} complements
the impossibility result for SOCO \cite{SOCO_OnlineOpt_UnreliablePrediction_Adam_Sigmetrics_2023_10.1145/3579442} by providing
the first condition under which finite robustness
and 1-consistency are simultaneously achievable.
 The intuition 
is that if the expert has a strictly positive
switching cost no less than $\epsilon>0$ at each time, 
then its per-step cost is also no less than $\epsilon$,
which provides \ouralg with the flexibility to choose a different action than the expert's action $x_t^{\pi}$ due to the $(1+\lambda)$ cost slackness in
the competitiveness requirement. Therefore, by choosing
a sufficiently large but still finite $\lambda\sim\mathcal{O}(\frac{1}{\epsilon})$, we
can show that $\Delta(\lambda)=0$ in \eqref{eqn:cr_bound_delay}
in Theorem~\ref{thm:setup_feedback_delay}, which means
\ouralg can completely follow the ML predictions.
Without this condition, it is possible that
the expert's cost is zero in the first few steps,
and hence \ouralg must follow the expert's actions at the beginning
to guarantee $(1+\lambda)$-competitiveness in case
the expert continues to have a zero cost in the future
--- even though
ML predictions are perfect and offline optimal, 
\ouralg cannot follow them at the beginning because
of the online process and $(1+\lambda)$-competitiveness requirement.
 
Importantly, our sufficient condition is not unrealistic in practice. For example, in moving object-tracking applications, the condition $d(x_{\tau}^\pi, x_{\tau-p:\tau-1}^\pi)\geq\epsilon>0$
is satisfied if the expert's action $\tilde{x}_t$ keeps changing to follow the moving object over time or alternatively, we ignore the dummy time steps with no movement.

\subsection{ML Model Training in \ouralg}\label{sec:single_nontrain}
We present the training details and highlight the advantage
of training the ML model in a \train manner to reduce
the average cost.

\subsubsection{Architecture, loss, and  dataset}\label{sec:ML_arch_loss_dataset}
Because of the recursive nature of SOCO and the strong representation power
of neural networks,
we use a recurrent neural network  with each base model parameterized
by $W\in\mathcal{W}$ (illustrated in Fig.~\ref{fig:illustration} in the appendix). Such architectures are also common in ML-based
optimizers for other online problems \cite{L2O_AdversarialOnlineResource_ChuanWu_HKU_TOMPECS_2021_10.1145/3494526,L2O_OnlineBipartiteMatching_Toronto_ArXiv_2021_DBLP:journals/corr/abs-2109-10380}.
With historical data available, we can construct
a training dataset $\mathcal{S}$ that contains a finite number
of problem instances.  The dataset can also be enlarged
using data augmentation techniques (e.g., adding adversarial samples) \cite{L2O_Adversarial_Robust_GAN_arXiv_2020,L2O_AdversarialOnlineResource_ChuanWu_HKU_TOMPECS_2021_10.1145/3494526}. 

\textbf{\Nontrain.} The existing literature on ML-augmented algorithms \cite{OnlineOpt_Learning_Augmented_RobustnessConsistency_NIPS_2020,SOCO_MetricUntrustedPrediction_Google_ICML_2020_pmlr-v119-antoniadis20a,SOCO_OnlineOpt_UntrustedPredictions_Switching_Adam_arXiv_2022,SOCO_ML_ChasingConvexBodiesFunction_Adam_COLT_2022} has commonly assumed that the ML model $h_W$ is 
separately trained 
in a \nontrain manner without being aware of the downstream
algorithm used online. Concretely,
the parameter $W$ of a \nontrain ML model is optimized for the following
loss
\begin{equation}\label{eqn:optimal_w_non_train}
\begin{split}
    W^*= \arg\min_{W\in\mathcal{W}}
  \frac{1}{|\mathcal{S}|}\sum_{\mathcal{S}}{cost}(\tilde{x}_{1:T}),
\end{split}
\end{equation}
where  $\tilde{x}_t = h_W\left(\tilde{x}_{t-p:t-1}, \left\{y_{\tau}|\tau\in\mathcal{D}_t^q\right\}\right)$ is the ML prediction
at time $t$.

\textbf{\Train.} 
There is a mismatch between the actual objective $\mathrm{cost}({x}_{1:T})$
and the training objective $\mathrm{cost}(\tilde{x}_{1:T})$ of a \nontrain ML model. To reconcile this, we propose to train the ML model
in a \train manner by explicitly considering the robustification
step in Algorithm~\ref{alg:RP-OBD_delay}.
For notational convenience, we denote the actual action as
 $x_t = \mathsf{Rob}_{\lambda}(h_W)=\mathsf{Rob}_{\lambda}(\tilde{x}_t)$, which
 emphasizes the projection of $\tilde{x}_t$
into the robust action set \eqref{eqn:reservation_hitting_delay}.
Thus, the parameter $W$ of  a \train ML model
is optimized to minimize the following loss
\begin{equation}\label{eqn:optimal_w_train}
 \hat{W}^*= \arg\min_{W\in\mathcal{W}}\frac{1}{|\mathcal{S}|}\sum_{\mathcal{S}}
{cost}(\mathsf{Rob}_{\lambda}(\tilde{x}_{1:T})), 
\end{equation}
which is different from \eqref{eqn:optimal_w_non_train} that only
minimizes the cost of pre-robustification ML predictions.

\subsubsection{Average cost} 
We bound the average
cost of \ouralg given an  ML model $h_W$.

\begin{theorem}[Average cost]\label{thm:average_non_train}
 Assume that the ML model is trained over a dataset $\mathcal{S}$ 
 drawn from the training distribution $\mathbb{P}'_y$.
 With probability at least $1-\delta, \delta\in(0,1)$, the average cost
 $\mathbb{E}_y\left[\mathrm{cost}(x_{1:T})\right]$ of 
 \ouralg over the testing distribution $y\sim\mathbb{P}_y$ is upper bounded by
\begin{equation}\label{eqn:generalization_avg_cost_non_train}
 \begin{split}
 \nonumber
 \overline{{cost}}(\ouralg)\leq\min
 \left\{(1+\lambda)\overline{{cost}}(\pi),
  \overline{{cost}}_{\mathcal{S}}(\mathsf{Rob}_{\lambda}(h_{W}))
 + \mathcal{O}\left(\mathrm{Rad}_{\mathcal{S}}(\mathsf{Rob}_{\lambda}(\mathcal{W}))+\sqrt{\frac{\log(\frac{2}{\delta})}{|\mathcal{S}|}}\right)  \right\},
\end{split}
\end{equation}
where $\overline{{cost}}_{\mathcal{S}}(\mathsf{Rob}_{\lambda}(h_{W}))=\frac{1}{|\mathcal{S}|}\sum_{\mathcal{S}}{cost}({x}_{1:T})$
is the empirical average cost  of 
robustified ML predictions in $\mathcal{S}$, 
 $\mathrm{Rad}_{\mathcal{S}}(\mathsf{Rob}_{\lambda}(\mathcal{W}))$ defined in Definition~\ref{definition:rademacher}
 in the appendix is the Rademacher complexity 
 with respect to the
 ML model space parameterized by $\mathcal{W}$ with robustification
 on the training dataset $\mathcal{S}$,
 the scaling coefficient inside $\mathcal{O}$
 for $\mathrm{Rad}_{\mathcal{S}}(\mathsf{Rob}_{\lambda}(\mathcal{W}))$
 is
 $\Gamma_x=\sqrt{T}|\mathcal{X}|\left[\beta_h+\frac{1}{2}(1+\sum_{i=1}^pL_i)(1+\sum_{i=1}^pL_i)\right]$ with $|\mathcal{X}|$ being the size of the action space $\mathcal{X}$ and $\beta_h$, $L_i$, and $p$ as  the smoothness constant, Lipschitz constant of the nonlinear term in the switching cost, and the memory length as defined in Assumptions \ref{assumption:hitting} and \ref{assumption:switching}, and 
 the coefficient for $\sqrt{\frac{\log(\frac{2}{\delta})}{|\mathcal{S}|}}$
 is $3\bar{c}$ with $\bar{c}$ being the upper bound of the total cost for an episode.
\end{theorem}

\Cref{thm:average_non_train} bounds the average cost of
\ouralg by
the minimum of two bounds. The first bound $(1+\lambda)\overline{{cost}}(\pi)$ further highlights
the guaranteed $(1+\lambda)$-competitiveness of \ouralg with respect to the expert's average
cost $AVG(\pi)$. 
The second bound includes a term 
$\overline{{cost}}_{\mathcal{S}}(\mathsf{Rob}_{\lambda}(h_{W}))=\frac{1}{|\mathcal{S}|}\sum_{\mathcal{S}}{cost}({x}_{1:T})$,
which is the empirical average cost of \ouralg given
an ML model $h_{W}$ and decreases when $\lambda>0$ increases.
The reason is that with a larger $\lambda>0$, the robust action set 
\eqref{eqn:proj_multi_delay_constraint} is enlarged and
\ouralg has more freedom to follow ML predictions
due to the less stringent competitiveness constraint.
Given an ML model $h_W$, Theorem~\ref{thm:setup_feedback_delay} shows how
the upper bound on $\overline{{cost}}_{\mathcal{S}}(\mathsf{Rob}_{\lambda}(h_{W}))$ varies with $\lambda$. Note that
 $h_{W^*}$ in \eqref{eqn:optimal_w_non_train} 
and $h_{\hat{W}^*}$ in \eqref{eqn:optimal_w_train}
minimize $\overline{{cost}}_{\mathcal{S}}(h_{{W}^*})$ and $\overline{{cost}}_{\mathcal{S}}(\mathsf{Rob}_{\lambda}(h_{\hat{W}^*}))$, respectively,
while the post-robustication cost (i.e., $\overline{{cost}}_{\mathcal{S}}(\mathsf{Rob}_{\lambda}(h_{\hat{W}^*}))$) is the actual cost of \ouralg.
Thus, we can further reduce the average cost by using
 the optimal \train ML model $h_{\hat{W}^*}$, compared to
 a \nontrain model $h_{{W}^*}$.

The other terms inside the second bound in \eqref{eqn:generalization_avg_cost_non_train}
are related to the training dataset: the larger dataset,
the smaller Rademacher complexity $\mathrm{Rad}_{\mathcal{S}}(\mathsf{Rob}(\mathcal{W}))$ and $\sqrt{\frac{\log(1/\delta)}{|\mathcal{S}|}}$.
Note that the Rademacher complexity $\mathrm{Rad}_{\mathcal{S}}(\mathsf{Rob}(\mathcal{W}))$ of
\ouralg is no greater than that of using the ML model alone (i.e. $\mathrm{Rad}_{\mathcal{S}}(\mathsf{Rob}(\mathcal{W}))\leq \mathrm{Rad}_{\mathcal{S}}\left(\mathcal{W}\right)$, shown in the appendix). The intuition is that \ouralg limits the action space for robustness. 
 Thus, Theorem~\ref{thm:setup_feedback_delay} provides
the insight that, given $\lambda>0$, robustification in \ouralg is 
more valuable in terms of bounding the average cost 
when the ML model $h_W$ is not well trained
(e.g., due to inadequate training data).
In practice, the hyperparameter
$\lambda>0$ can be set to improve
the empirical average performance
subject to the robustness constraint
based on a held-out validation dataset
along with the tuning of other hyperparameters (e.g., learning rates).

\subsubsection{\Train training and experimental verification}
Despite the advantage 
in terms of the average cost,
it is non-trivial to train a \train ML model using
standard back-propagation.
This is because
the operator of projecting ML predictions into a
robust set \eqref{eqn:proj_multi_delay_constraint}
is a recursive \emph{implicit} layer that cannot be 
easily differentiated as typical neural network layers.
Due to space limitations, we defer to the appendix the
the differentiation of the loss function with
respect to ML model weights $W$.

We validate the theoretical analysis by exploring the empirical performance of \ouralg using a case study of battery management in electric vehicle (EV) charging stations \cite{EV_Charging_Online_CityU_CUHK_TSG_2014}. %Due to space constraints, 
Our results are presented in the appendix. The results highlight the advantage
of \ouralg in terms of robustness compared to pure ML models, as well as the benefit of using a \train ML model in terms of the average cost.

%\subsection{Differentiable Expert Robustification}\label{sec:training}

\section{Experiments}\label{sec:experiment_main}

We now explore the performance of \ouralg using a case study focused on battery management in electric vehicle (EV) charging stations \citenewcite{EV_Charging_Online_CityU_CUHK_TSG_2014_dup}. 
We first formulate the problem as an instance of SOCO.
More details can be found at Appendix~\ref{sec:appendix_simulation_setup}.
Then, we test \ouralg on a public dataset provided by ElaadNL, compared with several
baseline algorithms including \robd\cite{SOCO_Memory_FeedbackDelay_Nonlinear_Adam_Sigmetrics_2022_10.1145/3508037}, \ecltwoo\cite{Shaolei_L2O_ExpertCalibrated_SOCO_SIGMETRICS_Journal_2022}, and \hitmin (details in Appendix~\ref{sec:baseline}). 
Our results highlight the advantage
of \ouralg in terms of robustness guarantees compared to pure ML models, as well
as the benefit of training a \train ML model in terms of the
average cost.

\subsection{Results}
We now present some results for
the case in which the hitting cost function (parameterized
by $y_t$) is immediately known without feedback delay. The results
for the case with feedback delay are presented in Appendix~\ref{sec:results_delay}.
Throughout the discussion, the reported values are normalized with respect to those of the respective \opt. The average cost (\textbf{AVG}) and competitive ratio (\textbf{CR}) are all empirical results reported on the testing dataset.

By Theorem~\ref{thm:setup_feedback_delay}, there is a trade-off (governed
by $\lambda>0$) between exploiting
ML predictions for good average performance and following the expert for robustness.  Here, we focus on the default setting of $\lambda=1$ and discuss the impact of different choices of $\lambda$ on both \ouralg and \ouralgnt in Appendix~\ref{sec:experiment_impact_lambda}. 
As shown in Table~\ref{table:default_setting}, with $\lambda=1$, both \ouralg and \ouralgnt have a good average cost,
but \ouralg has a lower average cost than \ouralgnt and is outperformed only by \ml in
terms of the average cost. \ouralg and \ouralgnt have the same competitive ratio
(i.e., $(1+\lambda)$ times the competitive ratio of \robd).
Empirically, \ouralg has the lowest competitive ratio among all the other algorithms, demonstrating the practical power of \ouralg for robustifying, potentially untrusted, ML predictions. In this experiment,
\ouralg outperforms \robd in terms of the empirical competitive ratio because
it exploits the good ML predictions for those problem instances that are adversarial to \robd. This result complements Theorem~\ref{thm:setup_feedback_delay}, where we show theoretically 
that \ouralg can outperform \robd in terms of the average cost by properly setting $\lambda$.
By comparison, \ml performs
well on average by exploiting the historical data, but 
has the highest competitive ratio due to its expected lack of robustness.
The two alternative baselines, \ecltwoo and \hitmin, are empirically good
on average and also in the worst case, but they do not have guaranteed robustness.
On the other hand, \robd is very robust, but its average cost is 
also the worst among all the algorithms under consideration. 

More results, including using \hitmin as the expert 
and large distributional shifts, are available in Appendix~\ref{sec:experiment_results}.

\begin{table}[!t]
\tiny\centering
\begin{tabular}{l|c|c|l|c|c|c|l|c|c|c|c|c} 
\toprule
& \multicolumn{4}{c|}{\textbf{\ouralg}}                    & \multicolumn{4}{c|}{\textbf{\ouralgnt}}               & \multirow{2}{*}{\textbf{\ml}} & \multirow{2}{*}{\textbf{\ecltwoo}} & \multirow{2}{*}{\textbf{\robd}} & \multirow{2}{*}{\textbf{\hitmin}}  \\ 
\cline{2-9}
& $\lambda$=0.6 & $\lambda$=1 & $\lambda$=3 & $\lambda$=5 & $\lambda$=0.6 & $\lambda$=1 & $\lambda$=3 & $\lambda$=5 & & & & \\ 
\hline
\multicolumn{1}{c|}{\textbf{AVG}} & 1.4704        & 1.1144      & 1.0531   & \textbf{1.0441}      & 1.4780        & 1.2432      &  1.0855   & 1.0738      & 1.0668                             & 1.1727                           & 1.6048                          & \multicolumn{1}{c}{1.2003}        \\ 
\hline
\multicolumn{1}{c|}{\textbf{CR}}  & 1.7672        & \textbf{1.2905}      &  1.4405  & 1.3014      & 2.2103        & 2.4209      & 2.4200   & 3.0322      & 3.2566                             & 2.0614                           & 1.7291                          & \multicolumn{1}{c}{2.0865}        \\
\cmidrule[\heavyrulewidth]{1-13}
\end{tabular}
\caption{Competitive ratio and average cost comparison of different algorithms.}\label{table:default_setting}
\end{table}

\section{Conclusion}

We have considered
a general SOCO setting (including multi-step switching costs
and delayed hitting costs) and proposed \ouralg, which ensures a worst-case performance bound
by utilizing an expert algorithm to robustify 
untrusted ML predictions.
We prove that \ouralg is able to guarantee 
$(1+\lambda)$-competitive against any given expert for any
$\lambda>0$. Additionally, 
we provide a sufficient condition that achieves
finite robustness and 1-consistency simultaneously.
To improve the average performance, we  explicitly train the ML model in a \train manner by differentiating
 the robustification step, and provide an explicit
 average cost  bound.
Finally,  we evaluate \ouralg using a case study of battery management for EV stations,
 which highlights the improvement in both robustness and average performance compared to existing algorithms. 

\textbf{Limitations and future work.}
We discuss two limitations of our work. First, we make
a few assumptions (e.g., convexity and smoothness) on the hitting costs
and switching costs. While they are common in the SOCO literature,
these assumptions may not always hold in practice, and relaxing them
will be interesting and challenging.
Second, for online optimization, we only consider the ML prediction for the current
step, whereas predictions for future steps can also be available in practice
and may be explored for performance improvement.
 There are also a number of interesting open questions that follow from this work. 
For example, it is  interesting to study alternative reservation costs and the optimality
 in terms of the tradeoff between bi-competitiveness.
Additionally, it would also be interesting to extend \ouralg to other related problems such
as convex body chasing and metrical task systems.

\textbf{Acknowledgement. } We would like to thank the anonymous reviewers for their helpful comments. Pengfei Li, Jianyi Yang and Shaolei Ren were supported in part by the U.S. NSF under the grant CNS--1910208. Adam Wierman was supported in part by the U.S. NSF under grants CNS-2146814, CPS-2136197, CNS-2106403, NGSDI-2105648.

\bibliographystyle{unsrt}
\bibliography{main_nips_final.bbl}
%\bibliography{ref_jy_knowledge,ref_pengfei,ref_ren}

\newpage
\appendix
\section{Illustration of \ouralg}

\begin{figure}[!h]	
\centering
	\includegraphics[width=0.82\textwidth]{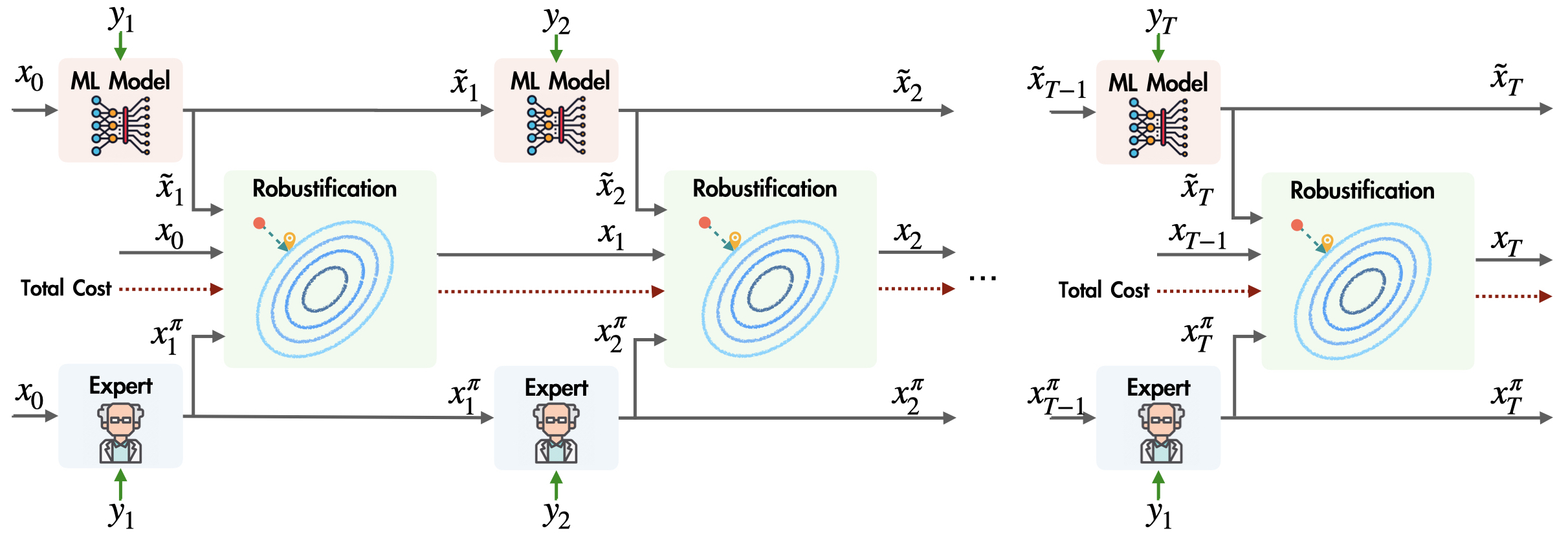}
	\vspace{-0.3cm}
	\caption{Robustness-constrained online optimization using \ouralg. The expert algorithm and ML model run independently. At each time $t=1,\cdots,T$, \ouralg %uses a robustification algorithm to
	projects the ML prediction $\tilde{x}_t$ into a robustified action set.} \label{fig:illustration}
\end{figure}

We illustrate the online optimization process of \ouralg
in Fig.~\ref{fig:illustration}.

\section{Case Study:  Battery Management for EV Charging Stations}\label{sec:experiment}

\subsection{Problem Formulation}\label{sec:appendix_simulation_setup}
Batteries are routinely used in EV charging stations to handle
the rapidly fluctuating charging demands and protect the connected
grid. Thus, properly 
managing battery charging/discharging decisions is crucial for
reliability, lifespan, and safety of batteries and grids.

We consider the management of $N$ batteries.
At each time step $t$, suppose that $x_t \in \mathbb{R}^N_+$ represents the State of Charge (SoC) and $u_t \in \mathbb{R}^N$ represents the battery charging/discharging  schedule, depending on the sign of $u_t$ (i.e., positive
means charging, and vice versa). The canonical form of the battery dynamics can be written as
  $  x_{t+1} = A x_t + B u_t - w_t$,
where $A$ is a $N\times N$ matrix which models the self-degradation of the $N$-battery system, $B$ is a $N\times N$ matrix which represents the charging efficiency of each battery unit, $w_t$ is a $N\times 1$ vector which denotes the current demand in terms of the charging rate (kW) of all the EVs connected to the charging stations. Assuming that
the initial SoC as $x_0$, the goal is to control the batteries
to minimize the 
difference between the current SoC of all batteries and a nominal value $\bar{x}$, plus a charging/discharging cost
to account for battery usage
 \citenewcite{SOCO_Memory_FeedbackDelay_Nonlinear_Adam_Sigmetrics_2022_10.1145/3508037_dup,ML_Control_MPC_Robust_Stability_TongxinLi_Adam_arXiv_2022}, which can be expressed mathematically as 
$\min_{u_1, u_2, \cdots, u_{T+1}} \sum_{t=1}^{T+1} \| x_t - \bar{x}\|^2 + b \| u_t\|^2$.

This problem falls into SOCO based on the reduction framework described in \citenewcite{SOCO_Memory_FeedbackDelay_Nonlinear_Adam_Sigmetrics_2022_10.1145/3508037_dup}. Specifically, at time step $t+1$, we can expand $x_{t+1}$ based on the battery dynamics
as  $ x_{t+1} = A^t x_1  + \sum_{j=1}^t A^{t-j}B u_j - \sum_{j=1}^t A^{t-j} w_j$,
We define the context parameter as $y_t = \bar{x} - A^t x_1 + \sum_{i=1}^t A^{t-i} w_i $ and the action as $a_t = \sum_{i=1}^t A^{t-i}B u_i$. 
 Then, assuming an identity matrix $B$ (ignoring charging loss), the optimization problem becomes
$\min_{a_1, \cdots, a_T} \| x_1 - \bar{x}\|^2 + b \| u_T\|^2 + \sum_{t=1}^{T} \| a_t - y_t \|^2 + b \| 
a_t - A  a_{t-1}\|^2$.
Given an initial value of $x_1$, 
this problem 
can be further simplified and reformulated as
\begin{equation}\label{eqn:cost_simulation}
    \min_{a_1, a_2, \cdots, a_{T}} \sum_{t=1}^{T} \frac{1}{b}\| a_t - y_t \|^2 +  \| 
    a_t - A  a_{t-1}\|^2,
\end{equation}
which is in a standard SOCO form by considering
$y_t$ as the context and $a_t$ as the action at time $t$.

To validate the effectiveness of \ouralg, we use a public dataset \citenewcite{develder2016quantifying} provided by ElaadNL, a Dutch EV charging infrastructure company. We collect
a dataset containing transaction records from ElaadNL charging stations in the Netherlands from January to June of 2019. Each transaction record contains the  energy demand, transaction start time and charging time. As the data does
not specify the details of battery units, we consider
the battery units as a single combine battery by summing up the energy demand
within each hour to obtain the hourly energy demand. 

We use the January to February data as the training dataset, March to April data as the validation dataset for tuning the hyperparameters such as learning rate, and May to June as the testing dataset. We consider each problem instance as
one day ($T=24$ hours, plus an initial action). Thus, a sliding window of 25 is applied, moving one hour ahead each time, on the raw data to generate 1416 problem instances, where the first demand of each instance is used as the initial action of all the algorithms. We set $b=10$ and $A=I$ for the cost function in Eqn.~\eqref{eqn:cost_simulation}.

All the algorithms use the same ML architecture, when applicable, with the same initialized weights 
in our experiments for fair comparison. 
To be consistent with the literature \citenewcite{L2O_OnlineBipartiteMatching_Toronto_ArXiv_2021_DBLP:journals/corr/abs-2109-10380_dup,L2O_AdversarialOnlineResource_ChuanWu_HKU_TOMPECS_2021_10.1145/3494526_dup}, all the ML models are trained offline. Specifically, we use a recurrent neural network (RNN) model that contains 2 hidden layers,
each with 8 neurons, and  implement the model using PyTorch. We train the RNN for 140 epochs with a batch size of 50. When the RNN model is trained 
as a standalone optimizer in a \nontrain manner, 
the training process takes around 1~minute on a 2020 MacBook Air with 8GB memory and a M1 chipset. When RNN is trained in a \train manner, it takes around 2~minutes. The testing process is almost instant and takes less than 1~second.

\subsection{Baseline Algorithms}\label{sec:baseline}

By default, \ouralg uses a \train ML model due to the advantage
of average cost performance compared to a \nontrain model.
We compare \ouralg with several representative baseline algorithms
as summarized below. 

$\bullet$ Offline Optimal Oracle (\textbf{\opt}): This is the optimal offline algorithm that has all the contextual information and optimally solves the problem.

$\bullet$ Regularized Online Balanced Descent (\textbf{\robd}):
\robd is the state-of-the-art order-optimal online algorithm with
the best-known competitive ratio for our SOCO setting \citenewcite{SOCO_OBD_R-OBD_Goel_Adam_NIPS_2019_NEURIPS2019_9f36407e_dup,SOCO_Memory_FeedbackDelay_Nonlinear_Adam_Sigmetrics_2022_10.1145/3508037_dup}. The parameters
of \robd are all optimally set according to \citenewcite{SOCO_Memory_FeedbackDelay_Nonlinear_Adam_Sigmetrics_2022_10.1145/3508037_dup}.
By default, \ouralg uses \robd as its expert for robustness.

$\bullet$ Hitting Cost Minimizer (\textbf{\hitmin}): \hitmin
is a special instance of \robd by setting the parameters such
that it greedily minimizing the hitting cost at each time.
This minimizer can be empirically effective and hence also used in 
\robd as a regularizer. 

$\bullet$ Machine Learning Only (\textbf{\ml}): \ml is trained as
a standalone optimizer in a \nontrain manner. It does not use
robustification during online optimization.

$\bullet$ Expert-Calibrated Learning (\textbf{\ecltwoo}): It is an ML-augmented algorithm that applies to our SOCO setting by using an ML model
to regularize online actions
 without robustness guarantees \citenewcite{Shaolei_L2O_ExpertCalibrated_SOCO_SIGMETRICS_Journal_2022_dup}. 
We set its parameters based on the validation dataset to have
the optimal average performance with an empirical competitive ratio less than $(1+\lambda)CR^{\pi}$.

$\bullet$ \ouralg with a \nontrain ML model (\textbf{\ouralgnt}):
To differentiate the two forms of \ouralg,
we use \ouralg to refer to \ouralg with
a \train ML model and \ouralgnt for the \nontrain ML model,
where ``$\texttt{-O}$'' represents  \nontrainnoun. 

To highlight our key contribution to the SOCO
literature,
the baseline algorithms we choose are representative of the state-of-the-art
expert algorithms, effective heuristics, and ML-augmented algorithms for the SOCO setting we consider. While there are a few other ML-augmented algorithms for SOCO \citenewcite{SOCO_MetricUntrustedPrediction_Google_ICML_2020_pmlr-v119-antoniadis20a_dup,SOCO_ML_ChasingConvexBodiesFunction_Adam_COLT_2022_dup,SOCO_OnlineOpt_UntrustedPredictions_Switching_Adam_arXiv_2022_dup}, they do not apply to our problem as they consider
unsquared switching costs in a metric space and
exploit the natural triangular inequality. 
Adapting them to the squared switching costs is non-trivial.

\subsection{Additional Empirical Results}\label{sec:experiment_results}
In Section~\ref{sec:experiment_main}, we have evaluated \ouralg using \robd as the expert online algorithm with $\lambda=1$. Here we provide additional experiment results to further evaluate the effectiveness of \ouralg, and the results are organized as follows. First, we change the expert online algorithm from \robd to \hitmin to show the flexibility of \ouralg. Second, we quantitatively present the effect of parameter $\lambda$ in \ouralg. Third, we introduce additional out-of-distribution samples to test the robustness of \ouralg and baselines, in terms of competitive ratio. Finally, we experiment under the delayed-feedback setting, which is a more challenging problem setup.

\begin{figure*}[t!]
	\centering
	\subfigure[\robd as the expert]{
		\includegraphics[width=0.23\textwidth]{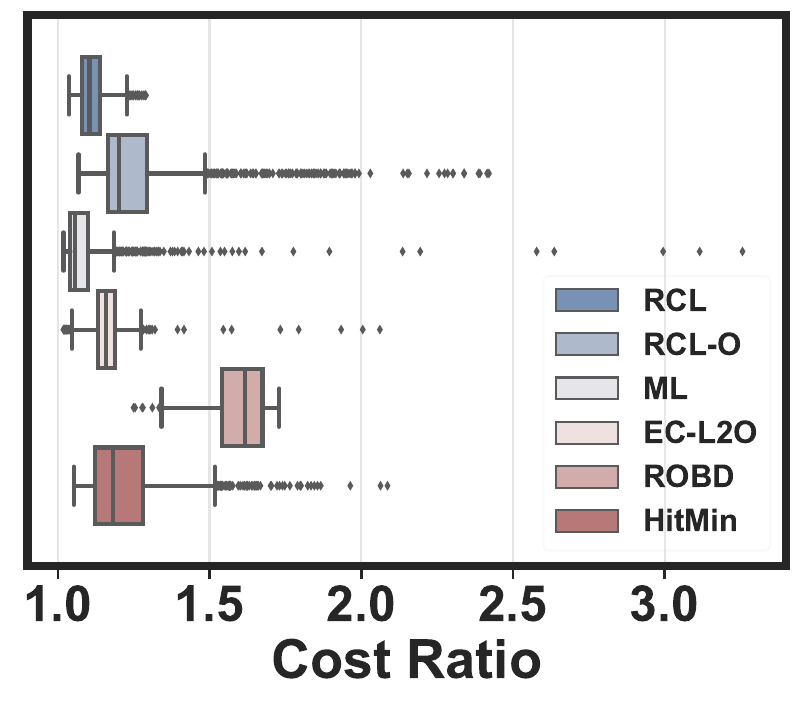}\label{fig:error_bar_follow_robd}
	}%	
	\hspace{0.015cm}
	\subfigure[\hitmin as the expert]{
		\includegraphics[width=0.23\textwidth]{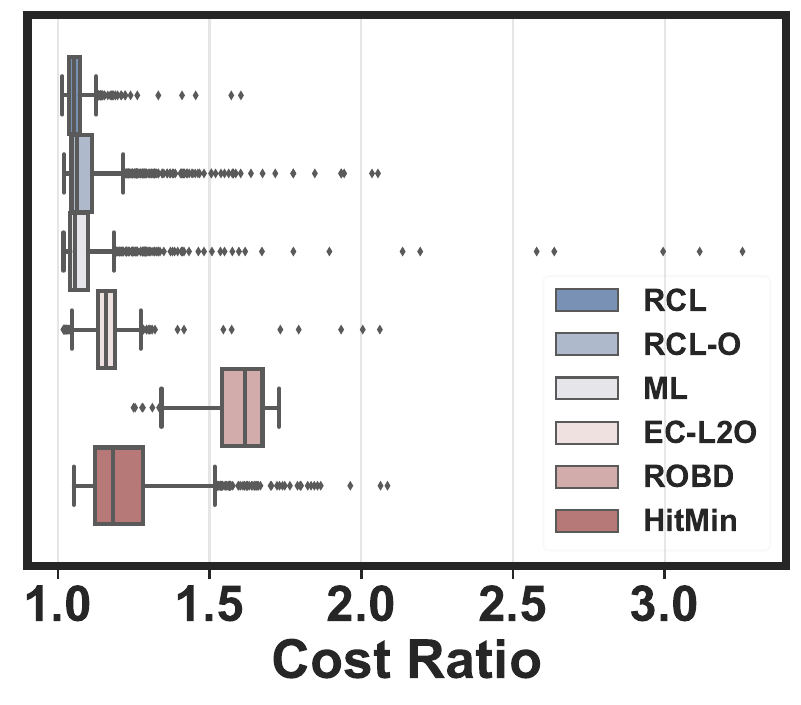}\label{fig:error_bar_follow_hitmin}
	}
	\subfigure[\ouralgnt w/ different $\lambda$]{
		\includegraphics[width=0.23\textwidth]{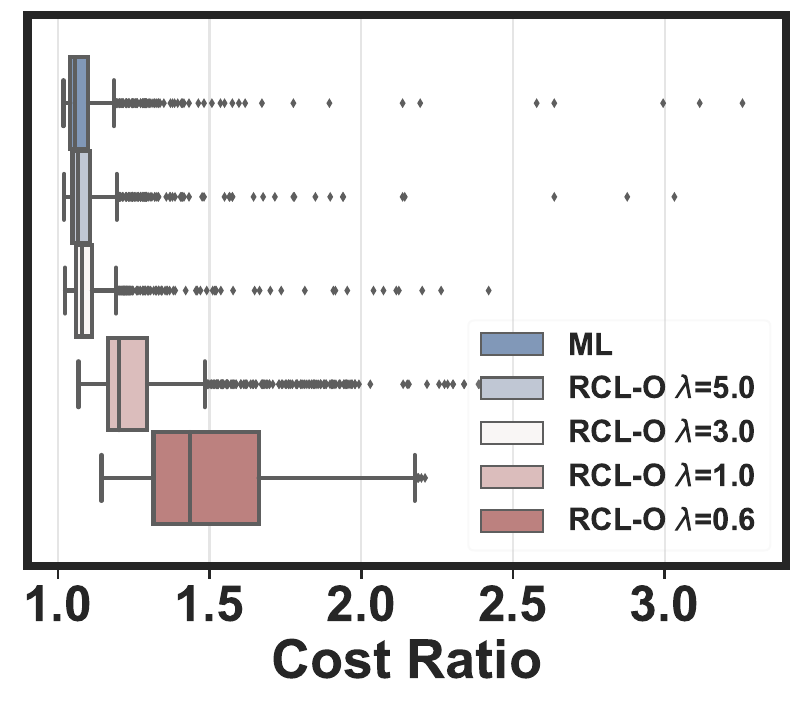}\label{fig:error_bar_elr_nt}
	}
	\subfigure[\ouralg w/ different $\lambda$]{
		\includegraphics[width=0.23\textwidth]{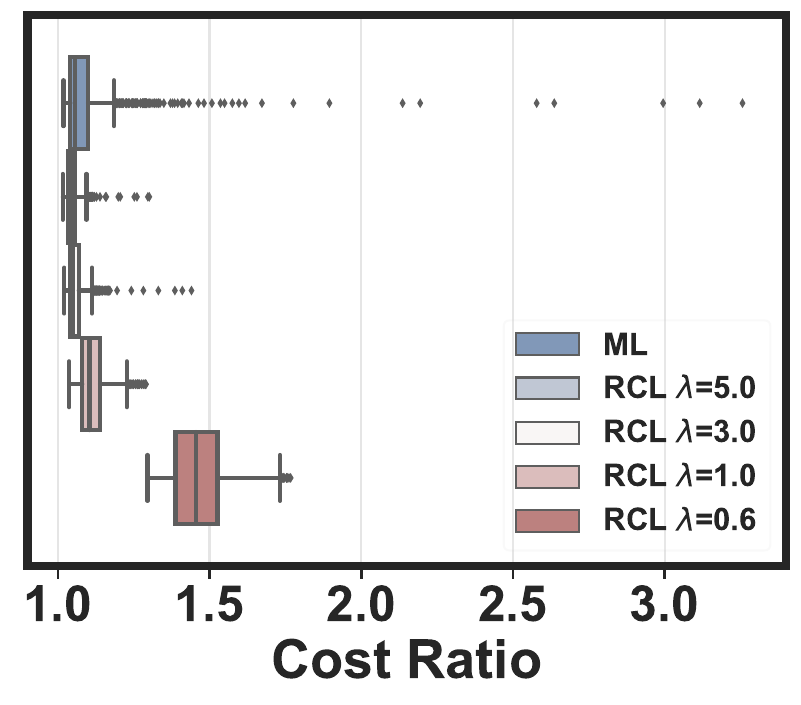}\label{fig:error_bar_erl_calib}
	}
 	\vspace{-0.3cm}	
	\caption{Cost ratio distributions ($\lambda = 1$ by default).}
 \label{fig:cost_ratio_all}
\end{figure*} 

\subsubsection{Utilizing \hitmin as the expert}
\ouralg is flexible and can work with any expert online algorithm,
even an expert that does not have good or bounded competitive ratios.
Thus, it is interesting to see how \ouralg performs given
an alternative expert.
For example, in Table~\ref{table:default_setting}, \hitmin empirically outperforms
\robd in terms of the average, although it is not as robust
 as \robd. 
Thus, using $\lambda=1$, we leverage \hitmin
 as the expert for \ouralg and \ouralgnt, and show
 the cost ratio distributions in Fig.~\ref{fig:error_bar_follow_hitmin}.
Comparing Fig.~\ref{fig:error_bar_follow_hitmin} with Fig.~\ref{fig:error_bar_follow_robd},
we see that \ouralg and \ouralgnt both have many low cost ratios by 
using \hitmin as the expert, but the worst case for \ouralg is not as
good as when using \robd as the expert.
For example, the average cost and competitive ratio are 1.0515 and 1.6035, respectively, for \ouralg. This result is not surprising, as the new expert \hitmin has a better average performance but worse competitive ratio than the default expert \robd.

\subsubsection{Impact of $\lambda$ }\label{sec:experiment_impact_lambda}
Theorem~\ref{thm:setup_feedback_delay}  shows the point that we need to set a large enough
$\lambda$ in order to provide enough flexibility for \ouralg to exploit
good ML predictions. With a small $\lambda>0$, despite
the stronger competitiveness against the expert, it is possible
that \ouralg may even empirically perform worse than both the ML model
and the expert. Thus, we now investigate the impact of $\lambda$.

We see from Table~\ref{table:default_setting} that the empirical
average cost and competitive ratio of \ouralg 
are both worse with $\lambda=0.6$ than with the default
$\lambda=1$. More interestingly, by setting $\lambda=5$, the average
cost of \ouralg is even lower than that of \ml. This is because \ml
in our experiment performs fairly well on average. Thus,
by setting a large $\lambda=5$, \ouralg is able to exploit
the benefits of good \ml predictions for many typical cases, while
using the expert \robd as a safeguard to handle a few bad
problem instances for which \ml cannot perform well.
Also,  the empirical competitive ratio of \ouralg is
 better with $\lambda=5$ than with $\lambda=3$, supporting
Theorem~\ref{thm:setup_feedback_delay}
that a larger $\lambda$ may not necessarily increase
the competitive ratio as \ouralg can exploit good ML predictions.
In addition, given each $\lambda$, \ouralg outperforms \ouralgnt,
which highlights the importance of training the ML model
in a \train manner to avoid the mismatch between training
and testing objectives.

We also show in Fig.~\ref{fig:error_bar_elr_nt} an Fig.~\ref{fig:error_bar_erl_calib} the cost ratio distributions
for \ouralgnt and \ouralg, respectively, under different $\lambda$.
The results reaffirm our main Theorem~\ref{thm:setup_feedback_delay} as well
as the importance of training the ML model in a \train manner.

\begin{figure*}[t]	
	\centering
	\subfigure[$\lambda=0.6$]{
	\includegraphics[width=0.23\textwidth]{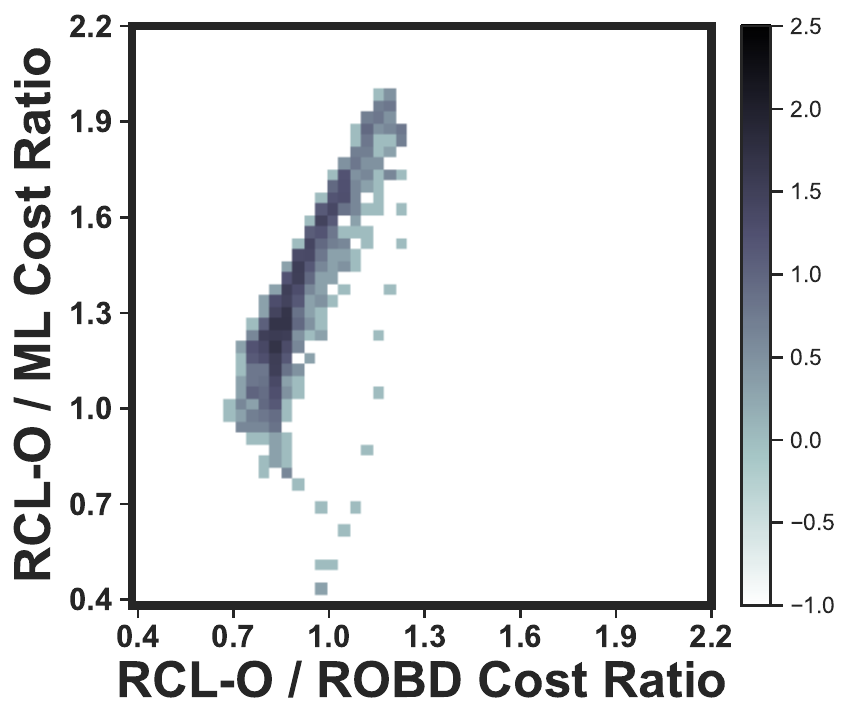}
	}%	
	\subfigure[$\lambda=1.0$]{
	\includegraphics[width=0.23\textwidth]{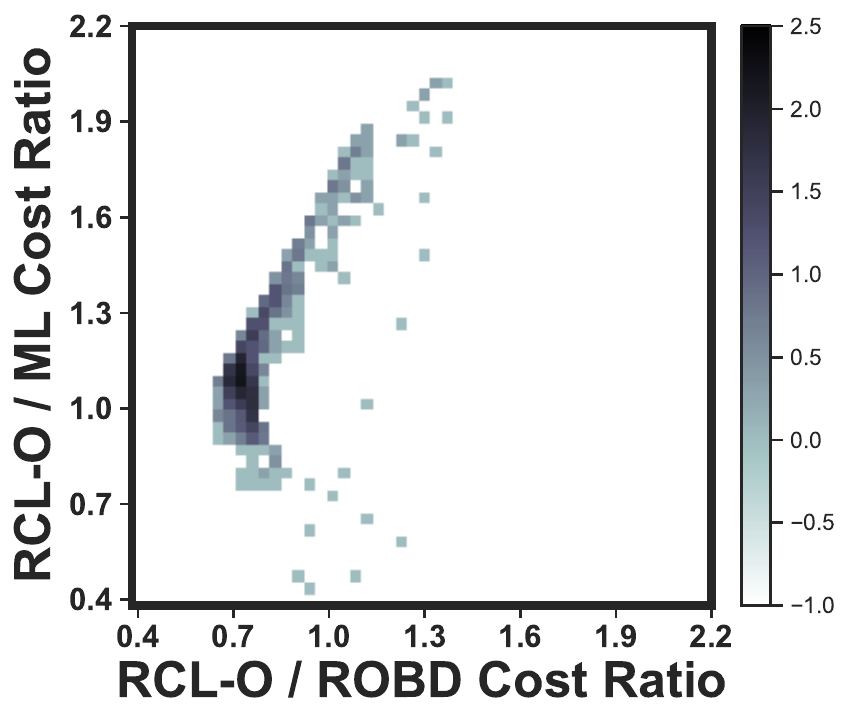}
	}
	\subfigure[$\lambda=5.0$]{
	\includegraphics[width=0.23\textwidth]{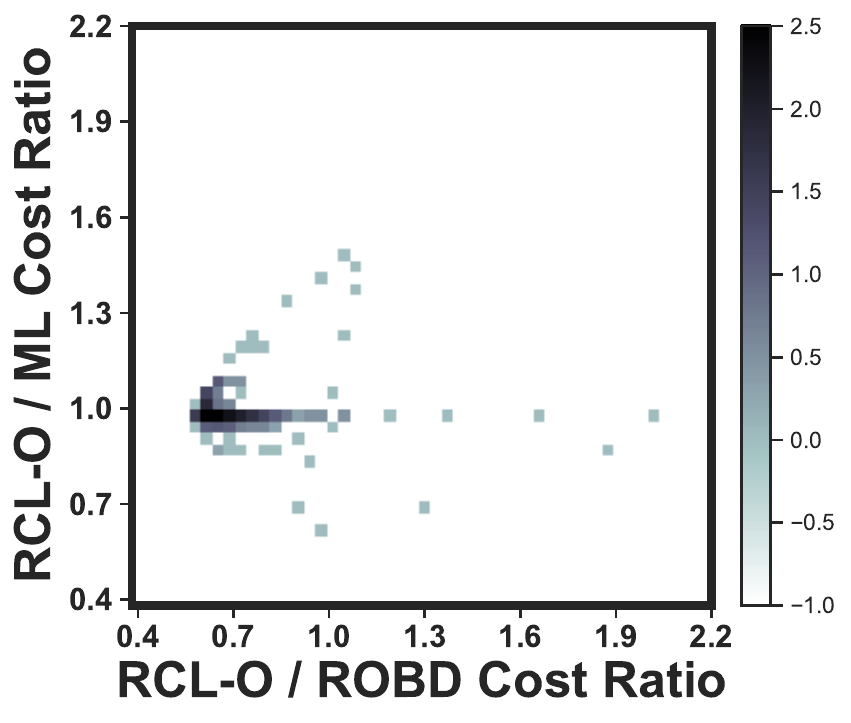}
	}
	\subfigure[$\lambda=\infty$ (i.e., \ml)]{
	\includegraphics[width=0.23\textwidth]{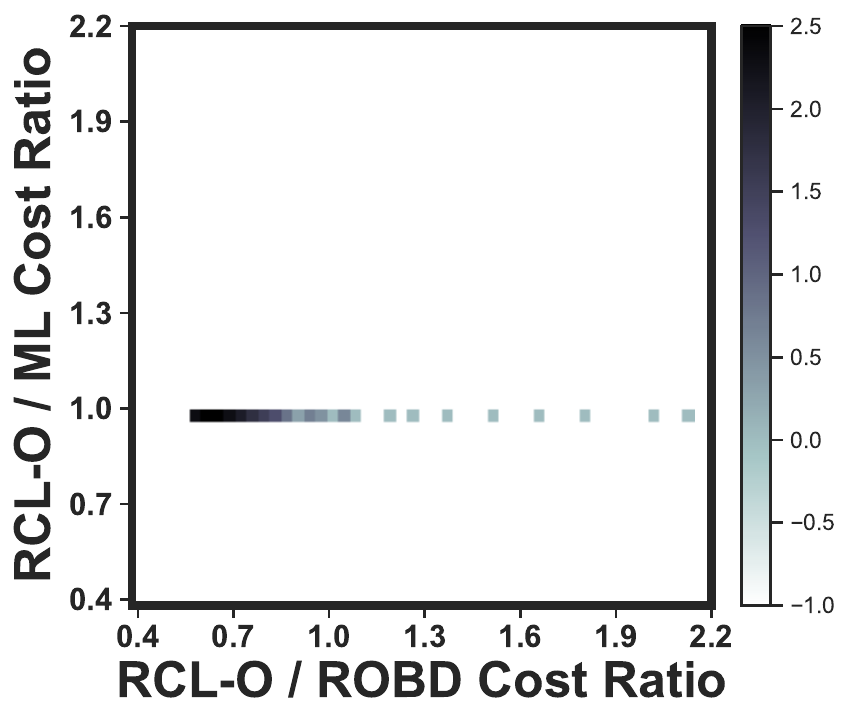}
	}
\vspace{-0.4cm}	
\caption{Histogram of bi-competitive cost ratios of \ouralgnt (against
\robd and \ml) under different $\lambda$. For better visualization, the color map represents
logarithmic values of the cost ratio histogram with a base of 10.}\label{fig:bi-competitiveness}
\end{figure*}

Next, we show the bi-competitive cost ratios of \ouralgnt against both the expert \robd and the \ml predictions. We focus on \ouralgnt as its ML model
is trained as a standalone optimizer, whereas \ouralg uses
a \train ML model that is not specifically trained
to produce good pre-robustification predictions.
According to Theorem~\ref{thm:setup_feedback_delay}, \ouralgnt
obtains a potentially better competitiveness
against \ml but a worse competitive against the expert \robd
when $\lambda$ increases, and vice versa.
To further validate the theoretical analysis, we test \ouralgnt with different $\lambda$ and obtain the 2D histogram of its bi-competitive cost ratios against
\robd and \ml, respectively. The results are  shown in Fig.~\ref{fig:bi-competitiveness}. 
In agreement with our analysis, the cost ratio of \ouralgnt against \robd  never exceeds $(1+\lambda)$ for any $\lambda>0$. Also, with
a small $\lambda=0.6$,  the cost ratio of \ouralgnt against \robd concentrates around 1, while it does not exploit the benefits of ML predictions very well.
On the other hand, with a large $\lambda=5$, the 
cost ratio of \ouralgnt  against \robd can be quite high,
although it follows (good) ML predictions more closely for better average performance. Most importantly, by increasing $\lambda>0$,
we can see the general trend that \ouralgnt follows
the \ml predictions more closely while still being
able to guarantee competitiveness against \robd. Again,
this confirms the key point of our main insights
in Theorem~\ref{thm:setup_feedback_delay}.

\begin{table}
 \centering
 \scriptsize
 \begin{tabular}{c|c|c|c|c|c|c|c|c|c|c} 
 \toprule
 \multicolumn{2}{c|}{\multirow{2}{*}{}} & \multicolumn{3}{c|}{\textbf{$p_c$=0.05}} & \multicolumn{3}{c|}{\textbf{$p_c$=0.1}} & \multicolumn{3}{c}{\textbf{$p_c$=0.2}} \\ 
 \cline{3-11}
 \multicolumn{2}{c|}{} & \textbf{$\sigma=$0.06} & \textbf{$\sigma=$0.08} & \textbf{$\sigma=$0.1} & \textbf{$\sigma=$0.06} & \textbf{$\sigma=$0.08} & \textbf{$\sigma=$0.1} & \textbf{$\sigma=$0.06} & \textbf{$\sigma=$0.08} & \textbf{$\sigma=$0.1} \\
 \hline
 \multirow{6}{*}{\textbf{AVG}} & \textbf{\ouralg} & 1.1331 & 1.1444 & 1.1556 & 1.1487 & 1.1693 & 1.1904 & 1.1827 & 1.2254 & 1.2697 \\ 
 \cline{2-11}
 & \textbf{\ouralgnt} & 1.2425 & 1.2436 & 1.2462 & 1.2416 & 1.2434 & 1.2478 & 1.2370 & 1.2394 & 1.2469 \\ 
 \cline{2-11}
 & \textbf{\ml} & {1.0722} & {1.0778} & {1.0855} & {1.0770} & {1.0874} & {1.1018} & {1.0858} & {1.1053} & {1.1325} \\ 
 \cline{2-11}
 & \textbf{\ecltwoo} & 1.1728 & 1.1737 & 1.1754 & 1.1731 & 1.1750 & 1.1784 & 1.1727 & 1.1757 & 1.1815 \\ 
 \cline{2-11}
 & \textbf{\robd} & 1.6048 & 1.6048 & 1.6048 & 1.6048 & 1.6049 & 1.6049 & 1.6048 & 1.6048 & 1.6049 \\ 
 \cline{2-11}
 & \textbf{\hitmin} & 1.2112 & 1.2195 & 1.2302 & 1.2202 & 1.2357 & 1.2557 & 1.2410 & 1.2724 & 1.3127 \\ 
 \specialrule{1pt}{0.5pt}{0.5pt}
 \multirow{6}{*}{\textbf{CR}} & \textbf{\ouralg} & 2.5028 & 2.9697 & 3.2247 & 2.6553 & 3.0283 & 3.2711 & 2.5714 & 3.0123 & 3.1653 \\ 
 \cline{2-11}
 & \textbf{\ouralgnt} & 2.4209 & 2.4209 & 2.4209 & 2.4209 & 2.4209 & 2.4209 & 2.4209 & 2.4209 & 2.4209 \\ 
 \cline{2-11}
 & \textbf{\ml} & 6.5159 & 8.9245 & 11.6627 & 4.4025 & 6.5090 & 9.4168 & 5.5798 & 7.3956 & 9.3903 \\ 
 \cline{2-11}
 & \textbf{\ecltwoo} & 3.4639 & 4.6034 & 5.9666 & 2.6545 & 3.6740 & 5.1129 & 2.9766 & 3.7713 & 4.6983 \\ 
 \cline{2-11}
 & \textbf{\robd} & {1.7291} & {1.7291} & {1.7291} & {1.7291} & {1.7291} & {1.7291} & {1.7291} & {1.7296} & {1.7298} \\ 
 \cline{2-11}
 & \textbf{\hitmin} & 4.8573 & 6.7746 & 8.8383 & 3.1492 & 4.8253 & 7.0405 & 5.0632 & 6.9699 & 8.9246 \\
 \bottomrule
 \end{tabular}
 \caption{Average cost and competitive ratio comparison
 of different algorithms. We study the effect of introducing
 out-of-distribution (OOD) samples. Within the testing dataset, we randomly select a fraction of $p_c$  of samples and add some random noise following $\mathcal{N}(0, \sigma)$ to contaminate these data samples (whose input
 values are all normalized within $[0,1]$).}\label{table:ood_results}
 \end{table}

\subsubsection{Larger distributional shifts}
In our dataset, \ml performs very well on average as
the testing distribution matches well with its training distribution.
To consider more challenging cases as a stress test, we manually increase
the testing distributional shifts by adding random noise following
$\mathcal{N}(0, \sigma)$
to a certain faction $p_c$ of the testing samples.
Note that, as we intentionally stress test
\ouralg and \ouralgnt under a larger distributional shift, 
their ML models remain unchanged as in the default setting
and are not re-trained by adding noisy data to the training dataset.

With the default $\lambda=1$, we show the average cost and competitive ratio results in Table~\ref{table:ood_results}. We
see that \robd is very robust and little affected by
the distributional shifts. In terms of the competitive ratio, \ml, \hitmin and \ecltwoo
are not robust, resulting in a large competitive
ratio when we add more noisy samples.
The average cost performance of \ouralg is empirically better
than that of \ouralgnt in almost all cases, except for a slight increase
in the practically very rare case
where 20\% samples are contaminated with large noise. 
On the other hand, as expected,
the competitive ratios of \ouralg and \ouralgnt
both increase as we add more noise.
While \ouralg has a higher competitive ratio than \ouralgnt empirically
in the experiment, they both have
the same guaranteed 
$(1+\lambda)$ competitiveness against \robd regardless of how their ML models
are trained. 
Also, their competitive ratios are both better than other algorithms, showing
 the effectiveness of our novel robustification process.

\subsection{Results with Feedback Delay}\label{sec:results_delay}

We now turn to the case when there is a one-step feedback delay, i.e.,
the context parameter $y_t$ is not known to the agent until time $t+1$. For
this setting, we consider the best-known online algorithm  \irobd  \citenewcite{SOCO_Memory_FeedbackDelay_Nonlinear_Adam_Sigmetrics_2022_10.1145/3508037_dup}
as the expert that handles the feedback delay with a guaranteed competitive ratio
with respect to OPT. The other baseline online algorithms --- \robd, \ecltwoo,
and \hitmin --- presented in Section~\ref{sec:baseline}
require the immediate revelation of $y_t$ without feedback delay and hence do not directly apply to this case. Thus, for comparison, we use the predicted context, denoted by $\hat{y}_t$, with up
to {15\%} prediction errors in the baseline online algorithms, and reuse
the algorithm names (e.g., \ecltwoo uses predicted $\hat{y}_t$ as if it
were the true context for decision making). We train \ml using
the same architecture as in Section~\ref{sec:experiment_results}, with the exception that only delayed context is provided as input for both training and testing. 
The reported values are normalized with respect to those of the respective offline optimal algorithm \opt. The average cost (\textbf{AVG}) and competitive ratio (\textbf{CR}) are all empirical results reported on the testing dataset.

\begin{figure*}[!t]	
	\centering
        \subfigure[\irobd as the expert]{
	\includegraphics[width=0.265\textwidth]{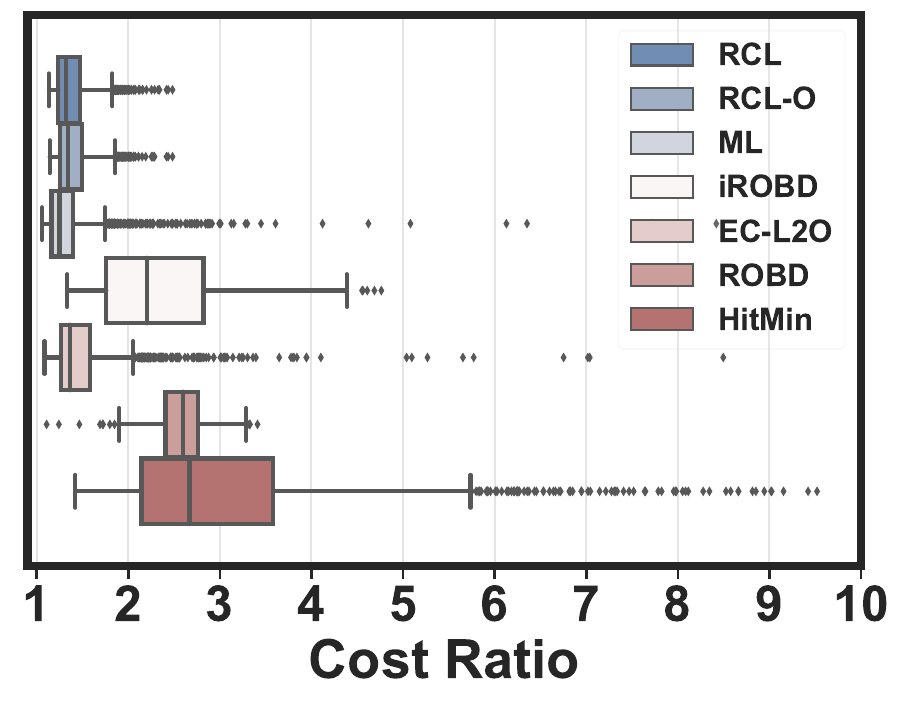}
	}
	\subfigure[\ouralg w/ different $\lambda$]{
	\includegraphics[width=0.26\textwidth]{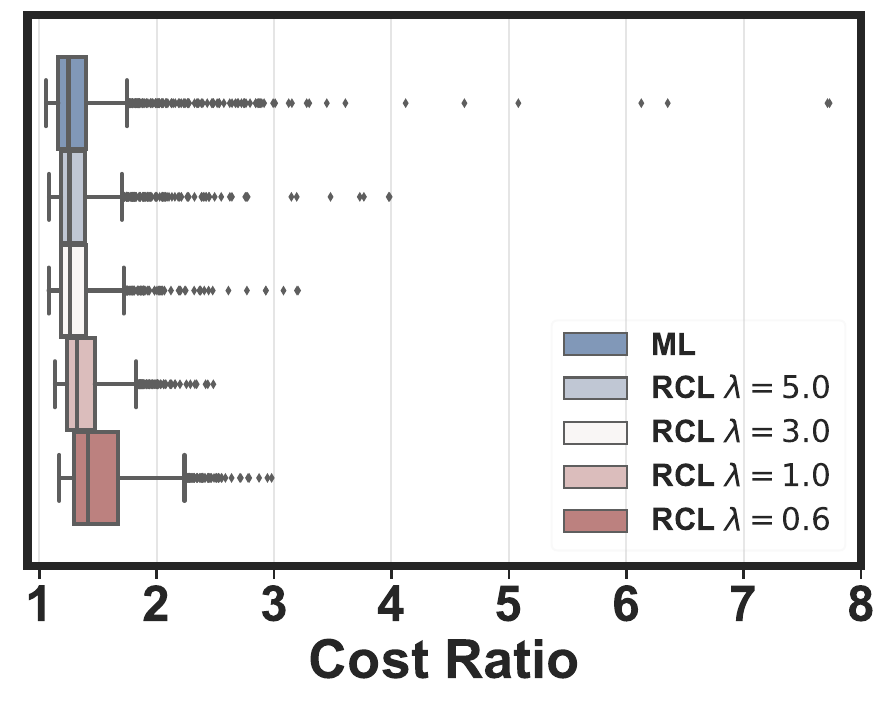}
	} 
	\subfigure[\ouralgnt w/ different $\lambda$]{
	\includegraphics[width=0.26\textwidth]{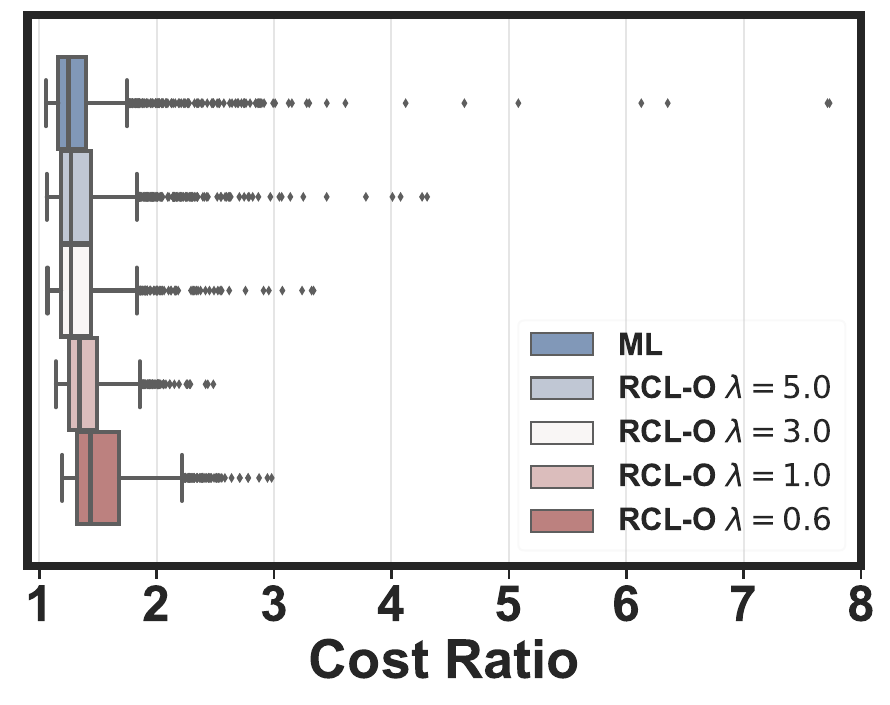}
	}
 \vspace{-0.3cm}
\caption{Cost ratio distributions with feedback delay ($\lambda=1$ by default)}\label{fig:cost_ratio_delay}
\end{figure*}

We show the results in Table~\ref{table:default_setting_delay} and Fig.~\ref{fig:cost_ratio_delay}.
We see that with the default $\lambda=1$, both \ouralg and \ouralgnt have a good average cost,
but \ouralg has a lower average cost than \ouralgnt and is outperformed only by \ml in
terms of the average cost. \ouralg and \ouralgnt have the same competitive ratio guarantee (i.e., $(1+\lambda)$ times the competitive ratio of \irobd).
Nonetheless, 
\ouralg has the lowest competitive ratio than all the other algorithms, demonstrating the power of \ouralg to leverage both ML prediction and the robust expert. 
In this experiment,
both \ouralg and \ouralgnt outperform \irobd in terms of the empirical competitive ratio because
they are able to exploit the good ML predictions for those problem instances that are difficult for \irobd. 

By comparison, \ml performs
well on average by exploiting the historical data, but 
has a high competitive ratio.
The alternative baselines --- \robd, \ecltwoo and \hitmin --- use
predicted context $\hat{y}_t$ as the true context. Except
for the good empirical competitive ratio of \robd,
they  do not have
good average performance or guaranteed robustness due to their
naively trusting the predicted context (that can potentially have large prediction errors). Note that the empirical competitive ratio of \robd with predicted
context is still much higher than that with the true context in Table~\ref{table:default_setting}.
These results reinforce the point
that blindly using ML predictions (i.e., predicted context in this example)
without additional robustification can lead to poor performance in terms of both
average cost and worst-case cost ratio.

We further show in Fig.~\ref{fig:cost_ratio_delay} the box plots for cost ratios 
of different algorithms, providing a detailed view of the algorithms' performance.
The key message is that  
\ouralg obtains the best of
both worlds --- a good average cost  
and a good competitive ratio. Moreover, we
see that by setting $\lambda=1$, we provide
enough freedom to \ouralg to exploit the benefits
of ML predictions while also ensuring worst-case robustness.
Thus, like in the no-delay case in Table~\ref{table:default_setting}
and Fig.~\ref{fig:cost_ratio_all}, the empirical competitive ratio of \ouralg with $\lambda=1$
is even lower than that with $\lambda=0.6$.

\begin{table}[!t]
\tiny\centering
\begin{tabular}{l|c|c|l|c|c|c|l|c|c|c|c|c|c} 
\toprule
& \multicolumn{4}{c|}{\textbf{\ouralg}}                    & \multicolumn{4}{c|}{\textbf{\ouralgnt}}               & \multirow{2}{*}{\textbf{\ml}} & \multirow{2}{*}{\textbf{\ecltwoo}} & \multirow{2}{*}{\textbf{iROBD}} & \multirow{2}{*}{\textbf{\hitmin}} &\multirow{2}{*}{\textbf{\robd}} \\ 
\cline{2-9}
& $\lambda$=0.6 & $\lambda$=1 & $\lambda$=3 & $\lambda$=5 & $\lambda$=0.6 & $\lambda$=1 & $\lambda$=3 & $\lambda$=5 & & & & \\ 
\hline
\multicolumn{1}{c|}{\textbf{AVG}} & 1.5011 & 1.3594 & 1.2874 & 1.2899 & 1.5134 & 1.3690 & 1.2949 & 1.3026 & \textbf{1.2792} & 1.4112 & 2.3076 & 2.6095   & 2.5974     \\ 
\hline
\multicolumn{1}{c|}{\textbf{CR}}  & 2.9797 & \textbf{2.4832} & 3.2049 & 3.9847 & 2.9797 & 2.4832 & 3.3367 & 4.3040 & 8.4200 & 15.1928 & 4.7632 & 26.0264   & 2.8478     \\
\cmidrule[\heavyrulewidth]{1-14}
\end{tabular}
\caption{Competitive ratio and average cost comparison of different algorithms with feedback delay.}\label{table:default_setting_delay}
\end{table}

\section{Proof of Theorems and Corollaries in  Section~\ref{sec:single_switching}}\label{appendix:proof_basic}

\subsection{Proof of Theorem~\ref{thm:setup_feedback_delay} (Cost Ratio)}\label{appendix:proof_single_steup}

To prove Theorem~\ref{thm:setup_feedback_delay}, we first give some technical lemmas about the smoothness of cost functions from Lemma~\ref{lemma:beta_triangle} to Lemma~\ref{lemma:total_smoothness_multi}. 
\begin{lemma}[Lemma 4 in \citenewcite{SOCO_Memory_Adam_NIPS_2020_NEURIPS2020_ed46558a_dup}]\label{lemma:beta_triangle}
Assume $f(x)$ is $\beta$ smooth, for any $\lambda >0$, we have 
$$f(x) \leq (1 + \lambda) f(y) + (1 + \frac{1}{\lambda})\frac{\beta}{2}\| x-y \|^2 \quad \forall x,y \in \mathcal{X}$$
\end{lemma}
\begin{lemma}\label{lemma:beta_add_together}
Assume $f(x)$ is $\beta_1$ smooth and $d(x)$ is $\beta_2$ smooth, then $f(x) + d(x)$ is $\beta_1 + \beta_2$ smooth.
\end{lemma}

\begin{lemma}\label{lemma:total_smoothness_multi}
Suppose that the hitting cost $f(x, y_t)$ is  $\beta_h$-smooth with respect to $x$, 
The switching cost is $d(x_t,x_{t-1})=\frac{1}{2} \|x_t- \delta(x_{t-p:t-1}) \|^2$, where $\delta(\cdot)$ is  $L_i$-Lipschitz with respect to $x_{t-i}$. Then for any two action sequences $x_{1:T}$ and $x_{1:T}'$, we must have
\begin{equation}
    \text{cost}(x_{1:T}) - (1+\lambda)\text{cost}(x_{1:T}') \leq \frac{\beta+ (1+ \sum_{k=1}^p L_k)^2}{2}(1+\frac{1}{\lambda})\|x_{1:T} - x_{1:T}'\|^2, \quad \forall \lambda >0
\end{equation}
\end{lemma}

\begin{proof}
The objective to be bounded can be decomposed as
\begin{equation}
    \begin{split}\label{eqn:multi_sum_hitting_smooth}
        &\text{cost}(x_{1:T}) - (1+\lambda)\text{cost}(x_{1:T}') \\
        =& \biggl(\sum_{t=1}^T f(x_t, y_t) - (1+\lambda)f(x_t', y_t) \biggr) + \frac{1}{2} \biggl( \sum_{t=1}^T \|x_t- \delta(x_{t-p:t-1})\|^2 - (1+\lambda)\|x_t'- \delta(x_{t-p:t-1}')\|^2 \biggr)
    \end{split}
\end{equation}
Since hitting cost is $\beta_h$-smooth, then
\begin{equation}\label{eqn:hitting_cost_bound_proof}
    \sum_{t=1}^T f(x_t, y_t) - (1+\lambda)f(x_t', y_t) \leq \frac{\beta_h}{2}(1+\frac{1}{\lambda}) \sum_{t=1}^T \|x_t - x_t' \|^2
\end{equation}
Besides, based on the Lipschitz assumption of function $\delta(\cdot)$, we have
\begin{equation}
\begin{split}
    &\|x_t- \delta(x_{t-p:t-1})\|^2 - (1+\lambda)\|x_t'- \delta(x_{t-p:t-1}')\|^2\\
    \leq& (1+\frac{1}{\lambda})\| (x_t - x_t') + (\delta(x_{t-p:t-1}) - \delta(x_{t-p:t-1}'))\|^2 \\
    \leq & (1+\frac{1}{\lambda})\left( \|x_t - x_t'\| + \|\delta(x_{t-p:t-1}) - \delta(x_{t-p:t-1}')\| \right)^2\\
    \leq & (1+\frac{1}{\lambda})\left( \|x_t - x_t'\| + \sum_{k=1}^p L_k \|x_{t-k} - x_{t-k}'\| \right)^2\\
    \leq & (1+\frac{1}{\lambda})(1 + \sum_{k=1}^p L_k)\left( \|x_t - x_t'\|^2 + \sum_{k=1}^p L_k \|x_{t-k} - x_{t-k}'\|^2 \right)
\end{split}
\end{equation}
Summing up the switching costs of all time steps together, we have
\begin{equation}
    \begin{split}\label{eqn:multi_sum_mem_smooth}
        & \sum_{t=1}^T  \|x_t- \delta(x_{t-p:t-1})\|^2 - (1+\lambda)\|x_t'- \delta(x_{t-p:t-1}')\|^2 \\
        \leq& (1+\frac{1}{\lambda})(1 + \sum_{k=1}^p L_k)\sum_{t=1}^T \left( \|x_t - x_t'\|^2 + \sum_{k=1}^p L_k \|x_{t-k} - x_{t-k}'\|^2 \right) \\
        \leq& (1+\frac{1}{\lambda})(1 + \sum_{k=1}^p L_k) \sum_{t=1}^T  (1 + \sum_{k=1}^p L_k)\|x_t - x_t'\|^2\\
        =& (1+\frac{1}{\lambda})(1 + \sum_{k=1}^p L_k)^2 \sum_{t=1}^T \|x_t - x_t'\|^2
    \end{split}
\end{equation}
Substituting Eqn.~\eqref{eqn:multi_sum_mem_smooth} and Eqn.~\eqref{eqn:hitting_cost_bound_proof} into Eqn.~\eqref{eqn:multi_sum_hitting_smooth}, we finish the proof.
\end{proof}

Now we propose Lemma \ref{lemma:cost_ratio_expert_delay} based on these above lemmas, which ensures the feasibility of robustness constraint in  Eqn.~\eqref{eqn:proj_multi_delay_constraint}
\begin{lemma}\label{lemma:cost_ratio_expert_delay}
Let $\pi$ be any expert algorithm for the SOCO problem with multi-step feedback delays and  multi-step switching costs, for any $\lambda \geq 0$ and $\lambda \geq \lambda_0 \geq 0$, the total cost by the projected actions $x_t$ must satisfy $\text{cost}(x_{1:T}) \leq (1+\lambda)\text{cost}(x_{1:T}^\pi)$
\end{lemma}

\begin{proof}
We prove by induction that the constraints in  Eqn.~\eqref{eqn:proj_multi_delay_constraint} are satisfied for each $t$.
For $t=1$, since we assume the initial actions are the same ($x_{-p+1:0} = x_{-p+1:0}^\pi$), it is obvious that $x = x_1^\pi$ satisfies the robustness constraints Eqn.~\eqref{eqn:proj_multi_delay_constraint}. 

Then for any time step $t \geq 2$, suppose it holds at $t-1$ that
\begin{equation}
\begin{aligned}\label{eqn:multi_delay_prev_constraint}
    &\sum_{\tau\in\mathcal{A}_{t-1}}f(x_{\tau}, y_\tau)+\sum_{\tau\in\mathcal{A}_{t-1} \cup \mathcal{B}_{t-1}}d(x_{\tau}, x_{\tau-p:\tau-1}) +\sum_{\tau\in\mathcal{B}_{t-1}}H(x_\tau, x_\tau^\pi)+ G(x, x_{t-p:t-1}, x_{t-p:t}^\pi)\\
    \leq&(1+\lambda)\biggl( \sum_{\tau\in\mathcal{A}_{t-1}}f(x^{\pi}_{\tau}, y_\tau) + \sum_{\tau\in\mathcal{A}_{t-1} \cup \mathcal{B}_{t-1}}d(x_{\tau}^\pi, x_{\tau-p:\tau-1}^\pi) \biggr)
\end{aligned}
\end{equation}
Now the robustness constraints Eqn.~\eqref{eqn:proj_multi_delay_constraint} is satisfied if we prove $x_t = x_t^\pi$ satisfies the constraints in Eqn.~\eqref{eqn:proj_multi_delay_constraint} at time step $t$. Since for the sets $\mathcal{A}$ and $\mathcal{B}$, we have
\begin{equation}
    \begin{split}
        (\mathcal{A}_{t} \cup \mathcal{B}_{t}) \backslash (\mathcal{A}_{t-1} \cup \mathcal{B}_{t-1}) = \{t\} ,\quad 
        \mathcal{A}_{t-1}\subseteq \mathcal{A}_{t},
    \end{split}
\end{equation}
so it holds that
\begin{equation}
    \sum_{\tau\in\mathcal{A}_{t} \cup \mathcal{B}_{t}}d(x_{\tau}, x_{\tau-p:\tau-1}) - \sum_{\tau\in\mathcal{A}_{t-1} \cup \mathcal{B}_{t-1}}d(x_{\tau}, x_{\tau-p:\tau-1}) = d(x_t, x_{t-p:t-1})
\end{equation}

By Lemma \ref{lemma:beta_triangle}, we have
\begin{equation}\label{eqn:multistepdistanceproof1}
    \begin{split}
        &d(x_t^\pi, x_{t-p:t-1}) - (1+\lambda)d(x_t^\pi, x_{t-p:t-1}^\pi) \\
        \leq & \frac{1}{2}(1+\frac{1}{\lambda}) \| \delta(x_{t-p:t-1}) - \delta(x_{t-p:t-1}^\pi) \|^2\\
        \leq &  \frac{1}{2}(1+\frac{1}{\lambda}) \left( \sum_{i=1}^p L_i \| x_{t-i} - x_{t-i}^\pi \| \right)^2
    \end{split}
\end{equation}

Denote $\alpha=1+ \sum_{k=1}^{p}L_k$. For the reservation cost, we have
\begin{equation}
    \begin{split}\label{eqn:multi_mem_reserve_diff}
        & G(x_{t-1}, x_{t-p-1:t-2}, x_{t-p-1:t-1}^\pi) - G(x_t^\pi, x_{t-p:t-1}, x_{t-p:t}^\pi) \\
        =& \frac{\alpha (1 + \frac{1}{\lambda_0}) }{2} \left(\sum_{k=1}^{p}  \sum_{i=0}^{p-k}L_{k+i} \| x_{t-i-1} - x_{t-i-1}^\pi\|^2 -  \sum_{k=1}^{p}  \sum_{i=1}^{p-k}L_{k+i} \| x_{t-i} - x_{t-i}^\pi\|^2 \right)  \\
        =& \frac{\alpha (1 + \frac{1}{\lambda_0}) }{2} \left(\sum_{k=0}^{p-1}  \sum_{i=1}^{p-k}L_{k+i} \| x_{t-i} - x_{t-i}^\pi\|^2 -  \sum_{k=1}^{p}  \sum_{i=1}^{p-k}L_{k+i} \| x_{t-i} - x_{t-i}^\pi\|^2 \right) \\
        =& \frac{\alpha (1 + \frac{1}{\lambda_0}) }{2} \left(\sum_{k=0}^{p-1}  \sum_{i=1}^{p-k}L_{k+i} \| x_{t-i} - x_{t-i}^\pi\|^2 -  \sum_{k=1}^{p-1} \sum_{i=1}^{p-k}L_{k+i} \| x_{t-i} - x_{t-i}^\pi\|^2 \right) \\
        =& \frac{\alpha (1 + \frac{1}{\lambda_0}) }{2} \sum_{i=1}^{p}L_{i} \| x_{t-i} - x_{t-i}^\pi\| ^2
    \end{split}
\end{equation}
Continuing with Eqn.~\eqref{eqn:multi_mem_reserve_diff}, we have
\begin{equation}
    \begin{split}
        &G(x_{t-1}, x_{t-p-1:t-2}, x_{t-p-1:t-1}^\pi) - G(x_t^\pi, x_{t-p:t-1}, x_{t-p:t}^\pi) =\frac{\alpha (1 + \frac{1}{\lambda_0}) }{2} \sum_{i=1}^{p}L_{i} \| x_{t-i} - x_{t-i}^\pi\| ^2 \\
        \geq & \frac{ (1 + \frac{1}{\lambda_0})(\sum_{i=1}^p L_i)^2  }{2} \sum_{i=1}^{p}\frac{L_{i}}{\sum_{i=1}^p L_i} \| x_{t-i} - x_{t-i}^\pi\| ^2\\
        \geq & \frac{ (1 + \frac{1}{\lambda_0})(\sum_{i=1}^p L_i)^2  }{2} \left( \sum_{i=1}^{p}\frac{L_{i}}{\sum_{i=1}^p L_i} \| x_{t-i} - x_{t-i}^\pi\| \right)^2 \\
        = & \frac{1}{2}(1+\frac{1}{\lambda_0}) \left( \sum_{i=1}^p L_i \| x_{t-i} - x_{t-i}^\pi \| \right)^2        \geq \frac{1}{2}(1+\frac{1}{\lambda}) \left( \sum_{i=1}^p L_i \| x_{t-i} - x_{t-i}^\pi \| \right)^2
    \end{split}
\end{equation}
where the second inequality holds by Jensen's inequality.
Therefore, combining with \eqref{eqn:multistepdistanceproof1}, we have
\begin{equation}\label{eqn:multi_mem_diff}
    d(x_t^\pi, x_{t-p:t-1}) + G(x_t^\pi, x_{t-p:t-1}, x_{t-p:t}^\pi) \leq G(x_{t-1}, x_{t-p-1:t-2}, x_{t-p-1:t-1}^\pi) + (1+\lambda)d(x_t^\pi, x_{t-p:t-1}^\pi)   
\end{equation}

By Eqn.~\eqref{eqn:multi_mem_diff}, we have
\begin{equation}
\begin{aligned}\label{eqn:multi_delay_switch_diff}
    &G(x_t^\pi, x_{t-p:t-1}, x_{t-p:t}^\pi) + \sum_{\tau\in\mathcal{A}_{t} \cup \mathcal{B}_{t}}d(x_{\tau}, x_{\tau-p:\tau-1}) - \sum_{\tau\in\mathcal{A}_{t-1} \cup \mathcal{B}_{t-1}}d(x_{\tau}, x_{\tau-p:\tau-1}) \\
    \leq& G(x_{t-1}, x_{t-p-1:t-2}, x_{t-p-1:t-1}^\pi) + (1+\lambda)\left( \sum_{\tau\in\mathcal{A}_{t} \cup \mathcal{B}_{t}}d(x_{\tau}^\pi, x_{\tau-p:\tau-1}^\pi) - \sum_{\tau\in\mathcal{A}_{t-1} \cup \mathcal{B}_{t-1}}d(x_{\tau}^\pi, x_{\tau-p:\tau-1}^\pi) \right)
\end{aligned}
\end{equation}
Now we define a new set $\mathcal{D}_{t} =  \mathcal{A}_{t}\backslash \mathcal{A}_{t-1}$, which denotes the timestep set for the newly received context parameters at $t$.

\textbf{Case 1}: If $t \in \mathcal{D}_{t}$, then $B_{t-1}\backslash B_{t} = \mathcal{D}_{t}\backslash \{t \}$, then we have
\begin{equation}
\begin{split}\label{eqn:multi_delay_hitting_case_2_eqn1}
    &\left(\sum_{\tau\in\mathcal{A}_{t}}f(x_{\tau}, y_\tau) +\sum_{\tau\in\mathcal{B}_{t}}H(x_\tau, x_\tau^\pi) \right) - \left(\sum_{\tau\in\mathcal{A}_{t-1}}f(x_{\tau}, y_\tau) +\sum_{\tau\in\mathcal{B}_{t-1}}H(x_\tau, x_\tau^\pi) \right)\\
    =& \sum_{\tau\in\mathcal{D}_{t}}f(x_{\tau}, y_\tau) - \sum_{\tau\in\mathcal{D}_{t}\backslash \{t\}}H(x_\tau, x_\tau^\pi) = f(x_t^\pi, y_t) + \sum_{\tau\in\mathcal{D}_{t}\backslash \{t\}}f(x_{\tau}, y_\tau) - \sum_{\tau\in\mathcal{D}_{t}\backslash \{t\}}H(x_\tau, x_\tau^\pi) 
\end{split}
\end{equation}
Since hitting cost $f(\cdot, y_t)$ is $\beta_h$-smooth, we have
\begin{equation}
\begin{split}\label{eqn:multi_delay_hitting_case_2_eqn2}
    &\sum_{\tau\in\mathcal{D}_{t}\backslash \{t\}}f(x_{\tau}, y_\tau) - \sum_{\tau\in\mathcal{D}_{t}\backslash \{t\}}(1+\lambda) f(x_{\tau}^\pi, y_{\tau}) \\
    \leq & \frac{\beta_h(1+\frac{1}{\lambda})}{2} \sum_{\tau\in\mathcal{D}_{t}\backslash \{t\}} \|x_{\tau}^\pi - x_{\tau} \|^2    \leq  \sum_{\tau\in\mathcal{D}_{t}\backslash \{t\}} H(x_\tau, x_\tau^\pi)
\end{split}
\end{equation}
Substituting Eqn.~\eqref{eqn:multi_delay_hitting_case_2_eqn2} back to Eqn.~\eqref{eqn:multi_delay_hitting_case_2_eqn1}, we have
\begin{equation}
\begin{split}\label{eqn:multi_delay_hitting_diff}
    &\left(\sum_{\tau\in\mathcal{A}_{t}}f(x_{\tau}, y_\tau) +\sum_{\tau\in\mathcal{B}_{t}}H(x_\tau, x_\tau^\pi) \right) - \left(\sum_{\tau\in\mathcal{A}_{t-1}}f(x_{\tau}, y_\tau) +\sum_{\tau\in\mathcal{B}_{t-1}}H(x_\tau, x_\tau^\pi) \right)\\
    \leq & (1+\lambda) \left(\sum_{\tau\in\mathcal{A}_{t}}f(x_{\tau}, y_\tau) - \sum_{\tau\in\mathcal{A}_{t-1}}f(x_{\tau}, y_\tau) \right)
\end{split}
\end{equation}

\textbf{Case 2}: If $t \notin \mathcal{D}_{t}$, then $(B_{t-1}\cup \{t \})\backslash B_{t} = \mathcal{D}_{t}$ and  we have
\begin{equation}
\begin{split}\label{eqn:multi_delay_hitting_case_1_eqn1}
    &\left(\sum_{\tau\in\mathcal{A}_{t}}f(x_{\tau}, y_\tau) +\sum_{\tau\in\mathcal{B}_{t}}H(x_\tau, x_\tau^\pi) \right) - \left(\sum_{\tau\in\mathcal{A}_{t-1}}f(x_{\tau}, y_\tau) +\sum_{\tau\in\mathcal{B}_{t-1}}H(x_\tau, x_\tau^\pi) \right)\\
    =& \sum_{\tau\in\mathcal{D}_{t}}f(x_{\tau}, y_\tau) - \sum_{\tau\in\mathcal{D}_{t}}H(x_\tau, x_\tau^\pi)+ H(x_t^\pi, x_t^\pi) \\
    =&  \sum_{\tau\in\mathcal{D}_{t} }f(x_{\tau}, y_\tau) - \sum_{\tau\in\mathcal{D}_{t} }H(x_\tau, x_\tau^\pi) 
\end{split}
\end{equation}
Since hitting cost $f(\cdot, y_t)$ is $\beta_h$-smooth, we have
\begin{equation}
\begin{split}\label{eqn:multi_delay_hitting_case_1_eqn2}
    &\sum_{\tau\in\mathcal{D}_{t} }f(x_{\tau}, y_\tau) - \sum_{\tau\in\mathcal{D}_{t} }(1+\lambda) f(x_{\tau}^\pi, y_{\tau}) \leq  \frac{\beta_h(1+\frac{1}{\lambda})}{2} \sum_{\tau\in\mathcal{D}_{t} } \|x_{\tau}^\pi - x_{\tau} \|^2 \leq  \sum_{\tau\in\mathcal{D}_{t} } H(x_\tau, x_\tau^\pi)
\end{split}
\end{equation}
Since $\lambda \geq 0$, we substitute Eqn.~\eqref{eqn:multi_delay_hitting_case_1_eqn2} back to Eqn.~\eqref{eqn:multi_delay_hitting_case_1_eqn1}, we have
the same conclusion as Eqn~\eqref{eqn:multi_delay_hitting_diff}.

Adding Eqn.~\eqref{eqn:multi_delay_prev_constraint}, Eqn.~\eqref{eqn:multi_delay_switch_diff} and Eqn.~\eqref{eqn:multi_delay_hitting_diff} together, we can prove $x = x_t^\pi$ satisfies the constraints in Eqn.~\eqref{eqn:proj_multi_delay_constraint}. At time step $T$, we have
\begin{equation}
\begin{split}
    &\sum_{\tau\in\mathcal{A}_T}f(x_{\tau}, y_\tau)+\sum_{\tau\in\mathcal{A}_T \cup \mathcal{B}_T}d(x_{\tau}, x_{\tau-p:\tau-1}) +\sum_{\tau\in\mathcal{B}_T}\left(f(x_{\tau}, y_\tau) - (1+\lambda)f(x_{\tau}^\pi, y_\tau) \right)\\
    \leq &\sum_{\tau\in\mathcal{A}_T}f(x_{\tau}, y_\tau)+\sum_{\tau\in\mathcal{A}_T \cup \mathcal{B}_T}d(x_{\tau}, x_{\tau-p:\tau-1}) +\sum_{\tau\in\mathcal{B}_T}H(x_\tau, x_\tau^\pi)\\
    \leq& (1+\lambda)\left( \sum_{\tau\in\mathcal{A}_T}f(x^{\pi}_{\tau}, y_\tau) + \sum_{\tau\in\mathcal{A}_T \cup \mathcal{B}_T}d(x_{\tau}^\pi, x_{\tau-p:\tau-1}^\pi) \right)
\end{split}
\end{equation}
In other words
\begin{equation}
\begin{split}
    \sum_{\tau\in\mathcal{A}_T \cup \mathcal{B}_T} \left(f(x_{\tau}, y_\tau) + d(x_{\tau}, x_{\tau-p:\tau-1}) \right)    \leq (1+\lambda)  \sum_{\tau\in\mathcal{A}_T \cup \mathcal{B}_T}\left(f(x^{\pi}_{\tau}, y_\tau) + d(x_{\tau}, x_{\tau-p:\tau-1}) \right)
\end{split}
\end{equation}

\end{proof}

In the next lemma, we bound the difference between the projected action and the ML predictions.

\begin{lemma} 
Suppose hitting cost is $\beta_h$-smooth, given the expert policy $\pi$, ML predictions $\tilde{x}_{1:T}$, for any $\lambda > 0$ and $\lambda_1 > 0$, the total distance between actual actions $x_{1:T}$ and ML predictions $\tilde{x}_{1:T}$ are bounded,
\begin{equation}
    \sum_{i=1}^T\|x_t - \tilde{x}_t \|^2 \leq  \sum_{i=1}^T \left( \left[\|\tilde{x}_t - x_t^\pi\| - \sqrt{K \left(d(x_t^\pi, x_{t-p:t-1}^\pi) + \sum_{\tau\in\mathcal{D}_{t}}f(x^{\pi}_{\tau}, y_\tau)\right) } \right]^+ \right)^2
\end{equation}
where $[\cdot]^+$ is the ReLU function and $K = \frac{2(\lambda - \lambda_0)}{\beta_h (1+\frac{1}{\lambda_0}) + \alpha^2 (1 + \frac{1}{\lambda_0})}$, $\alpha = 1+\sum_{i=1}^{p} L_i$
\end{lemma}
\begin{proof}
Suppose we at $t-1$ have the following inequality:
\begin{equation}
\begin{aligned}
    &\sum_{\tau\in\mathcal{A}_{t-1}}f(x_{\tau}, y_\tau)+\sum_{\tau\in\mathcal{A}_{t-1} \cup \mathcal{B}_{t-1}}d(x_{\tau}, x_{\tau-p:\tau-1}) +\sum_{\tau\in\mathcal{B}_{t-1}}H(x_\tau, x_\tau^\pi)+ G(x, x_{t-p:t-1}, x_{t-p:t}^\pi)\\
    \leq& (1+\lambda)\left( \sum_{\tau\in\mathcal{A}_{t-1}}f(x^{\pi}_{\tau}, y_\tau) + \sum_{\tau\in\mathcal{A}_{t-1} \cup \mathcal{B}_{t-1}}d(x_{\tau}^\pi, x_{\tau-p:\tau-1}^\pi) \right)
\end{aligned}
\end{equation}
Remember that $\mathcal{D}_{t} =  \mathcal{A}_{t}\backslash \mathcal{A}_{t-1}$ is the set of the time steps for the newly received context parameters at $t$. The robustness constraint in Eqn.~\eqref{eqn:proj_multi_delay_constraint} is satisfied if $x_t$ satisfies the following inequality.

\begin{equation}
\begin{aligned}\label{eqn:multi_feedback_delay_eqn1}
    &\left(\sum_{\tau\in\mathcal{D}_{t}}f(x_{\tau}, y_\tau) +\sum_{\tau\in\mathcal{B}_{t}}H(x_\tau, x_\tau^\pi) - \sum_{\tau\in\mathcal{B}_{t-1}}H(x_\tau, x_\tau^\pi) \right) + d(x_t, x_{t-p:t-1}) + G(x_t, x_{t-p:t-1}, x_{t-p:t}^\pi)\\
    -& G(x_{t-1}, x_{t-p-1:t-2}, x_{t-p-1:t-1}^\pi) \leq(1+\lambda)\left(d(x_{t}^\pi, x_{t-p:t-1}^\pi) +  \sum_{\tau\in\mathcal{D}_{t}}f(x^{\pi}_{\tau}, y_\tau)  \right)
\end{aligned}
\end{equation}

For the switching cost, we have
\begin{equation}\label{eqn:memorycostboundmulti}
    \begin{split}
        &d(x, x_{t-p:t-1}) - (1+\lambda_0)d(x_t^\pi, x_{t-p:t-1}^\pi) \\
        \leq & \frac{1}{2}(1+\frac{1}{\lambda_0})\left( \|x - x_t^\pi \| +  \| \delta(x_{t-p:t-1}) - \delta(x_{t-p:t-1}^\pi) \| \right)^2\\
        \leq &  \frac{1}{2}(1+\frac{1}{\lambda_0}) \left( \|x - x_t^\pi \| + \sum_{i=1}^p L_i \| x_{t-i} - x_{t-i}^\pi \| \right)^2\\
        \leq & \frac{ \alpha (1 + \frac{1}{\lambda_0}) }{2} \left( \|x - x_t^\pi \|^2 + \sum_{i=1}^p L_i \| x_{t-i} - x_{t-i}^\pi \|^2 \right)
    \end{split}
\end{equation}
The first inequality comes from Lemma \ref{lemma:beta_triangle}, the second inequality comes from the $L_i$-Lipschitz assumption, and the third inequality is because $\alpha\geq 1$. Besides, from Eqn \eqref{eqn:multi_mem_reserve_diff}, we have
\begin{equation}
    \begin{split}
        G(x_{t-1}, x_{t-p-1:t-2}, x_{t-p-1:t-1}^\pi) - G(x_t^\pi, x_{t-p:t-1}, x_{t-p:t}^\pi)  = \frac{\alpha (1 + \frac{1}{\lambda_0}) }{2} \sum_{i=1}^{p}L_{i} \| x_{t-i} - x_{t-i}^\pi\| ^2
    \end{split}
\end{equation}
Thus we have
\begin{equation}
    \begin{split}
        &G(x, x_{t-p:t-1}, x_{t-p:t}^\pi) - G(x_{t-1}, x_{t-p-1:t-2}, x_{t-p-1:t-1}^\pi)\\
        =& G(x, x_{t-p:t-1}, x_{t-p:t}^\pi) - G(x_t^\pi, x_{t-p:t-1}, x_{t-p:t}^\pi) + G(x_t^\pi, x_{t-p:t-1}, x_{t-p:t}^\pi) - G(x_{t-1}, x_{t-p-1:t-2}, x_{t-p-1:t-1}^\pi)\\
        =&G(x, x_{t-p:t-1}, x_{t-p:t}^\pi) - G(x_t^\pi, x_{t-p:t-1}, x_{t-p:t}^\pi) - \frac{\alpha (1 + \frac{1}{\lambda_0}) }{2} \sum_{i=1}^{p}L_{i} \| x_{t-i} - x_{t-i}^\pi\| ^2.
    \end{split}
\end{equation}
Combining with inequality \eqref{eqn:memorycostboundmulti}, we have
\begin{equation}
    \begin{split}\label{eqn:multi_step_mem_distance_bound}
         &G(x_t, x_{t-p:t-1}, x_{t-p:t}^\pi) - G(x_{t-1}, x_{t-p-1:t-2}, x_{t-p-1:t-1}^\pi) + d(x_t, x_{t-p:t-1}) - (1+\lambda_0)d(x_t^\pi, x_{t-p:t-1}^\pi)\\
         \leq & G(x_t, x_{t-p:t-1}, x_{t-p:t}^\pi) - G(x_t^\pi, x_{t-p:t-1}, x_{t-p:t}^\pi) + \frac{ \alpha (1 + \frac{1}{\lambda_0}) }{2}  \|x_t - x_t^\pi \|^2\\
         = & \frac{\alpha (1 + \frac{1}{\lambda_0}) \sum_{k=1}^{p}  L_k}{2}  \| x_t - x_t^\pi\|^2  + \frac{\alpha (1 + \frac{1}{\lambda_0}) }{2} \sum_{k=1}^{p} \| x_t - x_t^\pi\|^2\\
         = &\frac{\alpha^2 (1 + \frac{1}{\lambda_0}) }{2}  \| x_t - x_t^\pi\|^2
    \end{split}
\end{equation}

Substituting Eqn.~\eqref{eqn:multi_step_mem_distance_bound} back to Eqn.~\eqref{eqn:multi_feedback_delay_eqn1}, we have
\begin{equation}
\begin{split}\label{eqn:multi_feedback_delay_eqn2}
    &\sum_{\tau\in\mathcal{D}_{t}}\left(f(x_{\tau}, y_\tau) - (1+\lambda_0)f(x_{\tau}^\pi, y_\tau) \right)+\sum_{\tau\in\mathcal{B}_{t}}H(x_\tau, x_\tau^\pi) - \sum_{\tau\in\mathcal{B}_{t-1}}H(x_\tau, x_\tau^\pi)  \\
    &+ \frac{\alpha^2 (1 + \frac{1}{\lambda_0}) }{2}  \| x - x_t^\pi\|^2 \leq(\lambda-\lambda_0)\left(d(x_{t}^\pi, x_{t-p:t-1}^\pi) +  \sum_{\tau\in\mathcal{D}_{t}}f(x^{\pi}_{\tau}, y_\tau) \right)
\end{split}
\end{equation}

\textbf{Case 1}: If $t \in \mathcal{D}_{t}$, then $B_{t-1}\backslash B_{t} = \mathcal{D}_{t}\backslash \{t \}$, then Eqn.\eqref{eqn:multi_feedback_delay_eqn2} becomes
\begin{equation}
\begin{split}
    &f(x_t, y_t) - (1+\lambda_0)f(x_t^\pi, y_t) + \frac{\alpha^2 (1 + \frac{1}{\lambda_0}) }{2}  \| x - x_t^\pi\|^2\\
    + \sum_{\tau\in\mathcal{D}_{t} \backslash \{t \} }f(x_{\tau}, y_\tau) - &(1+\lambda_0)f(x_{\tau}^\pi, y_\tau) - H(x_\tau, x_\tau^\pi)   \leq(\lambda-\lambda_0)\left(d(x_{t}^\pi, x_{t-p:t-1}^\pi) +  \sum_{\tau\in\mathcal{D}_{t}}f(x^{\pi}_{\tau}, y_\tau) \right)
\end{split}
\end{equation}
Since hitting cost is $\beta_h$-smooth, the sufficient condition for Eqn.~\eqref{eqn:multi_feedback_delay_eqn2} becomes
\begin{equation}
    \frac{(\beta_h+ \alpha^2) (1 + \frac{1}{\lambda_0}) }{2}  \| x - x_t^\pi\|^2 \leq  (\lambda-\lambda_0)\left(d(x_{t}^\pi, x_{t-p:t-1}^\pi) +  \sum_{\tau\in\mathcal{D}_{t}}f(x^{\pi}_{\tau}, y_\tau) \right)
\end{equation}
Since the hitting cost is non-negative, the sufficient condition can be further simplified, which is
\begin{equation}
    \frac{(\beta_h+ \alpha^2) (1 + \frac{1}{\lambda_0}) }{2}  \| x - x_t^\pi\|^2 \leq  (\lambda-\lambda_0)\left( f(x^{\pi}_t, y_t) + d(x_{t}^\pi, x_{t-p:t-1}^\pi) \right)
\end{equation}

\textbf{Case 2}: If $t \notin \mathcal{D}_{t}$, then $(B_{t-1}\cup \{t \})\backslash B_{t} = \mathcal{D}_{t}$, then Eqn.\eqref{eqn:multi_feedback_delay_eqn2} becomes
\begin{equation}
\begin{split}\label{eqn:multi_feedback_delay_eqn3}
    &\frac{\alpha^2 (1 + \frac{1}{\lambda_0}) }{2}  \| x - x_t^\pi\|^2  + H(x, x_t^\pi) + \sum_{\tau\in\mathcal{D}_{t}}\left(f(x_{\tau}, y_\tau) - (1+\lambda_0)f(x_{\tau}^\pi, y_\tau) - H(x_\tau, x_\tau^\pi) \right)   \\
    & \leq(\lambda-\lambda_0)\left(d(x_{t}^\pi, x_{t-p:t-1}^\pi) +  \sum_{\tau\in\mathcal{D}_{t}}f(x^{\pi}_{\tau}, y_\tau) \right)
\end{split}
\end{equation}
Since hitting cost is $\beta_h$-smooth, the sufficient condition for Eqn.~\eqref{eqn:multi_feedback_delay_eqn3} becomes
\begin{equation}
    \frac{(\beta_h+ \alpha^2) (1 + \frac{1}{\lambda_0}) }{2}  \| x - x_t^\pi\|^2 \leq  (\lambda-\lambda_0)\left(d(x_{t}^\pi, x_{t-p:t-1}^\pi) +  \sum_{\tau\in\mathcal{D}_{t}}f(x^{\pi}_{\tau}, y_\tau) \right)
\end{equation}

Now we define 
$$K = \frac{2(\lambda - \lambda_0)}{(\beta_h+ \alpha^2) (1 + \frac{1}{\lambda_0})}$$

At time step $t$, if $x_t'$ is the solution to this alternative optimization problem
\begin{equation}
    \begin{split}
        &x_t' = \text{arg} \min_{x} \frac{1}{2} \|x - \tilde{x}_t \|^2 \\
        s.t. \quad & \|x - x_t^\pi \|^2 \leq K \left(d(x_t^\pi, x_{t-p:t-1}^\pi) + \sum_{\tau\in\mathcal{D}_{t}}f(x^{\pi}_{\tau}, y_\tau)\right)
    \end{split}
\end{equation}
The solution to this problem can be calculated asd
\begin{equation}
    \begin{split}
        x_t' &= \theta x_t^\pi + (1-\theta) \tilde{x}_t\\
        \theta&= \left[ 1- \frac{\sqrt{K \left(d(x_t^\pi, x_{t-p:t-1}^\pi) + \sum_{\tau\in\mathcal{D}_{t}}f(x^{\pi}_{\tau}, y_\tau)\right) }}{\| \tilde{x}_t -  x_t^\pi\|}\right]^+.
    \end{split}
\end{equation}
Then $\|x_t' - \tilde{x}_t \| = \left[\|\tilde{x}_t - x_t^\pi\| - \sqrt{K \left(d(x_t^\pi, x_{t-p:t-1}^\pi) + \sum_{\tau\in\mathcal{D}_{t}}f(x^{\pi}_{\tau}, y_\tau)\right) } \right]^+$. Since $x_t'$ also satisfies the original robustness constraint, we have $\|x_t - \tilde{x}_t \| \leq \|x_t' - \tilde{x}_t \|$ and we finish the proof.

\end{proof}

\noindent\textbf{Proof of Theorem \ref{thm:setup_feedback_delay}}\\ 
Now summing up the distance through 1 to T, we have
\begin{equation}
    \sum_{i=1}^T\|x_t - \tilde{x}_t \|^2 \leq  \sum_{i=1}^T \left( \left[\| \tilde{x}_t  - x_t^\pi\| - \sqrt{K \left(d(x_t^\pi, x_{t-p:t-1}^\pi) + \sum_{\tau\in\mathcal{D}_{t}}f(x^{\pi}_{\tau}, y_\tau) \right)}\right]^+ \right)^2
\end{equation}
Based on Lemma \ref{lemma:total_smoothness_multi} we have $\forall \lambda_2 >0$,    
\begin{equation}\label{eqn:consistency_cost_delay}
    \text{cost}(x_{1:T}) - (1+\lambda_2)\text{cost}(\tilde{x}_{1:T}) \leq \frac{\beta+ \alpha^2}{2}(1+\frac{1}{\lambda_2})\sum_{i=1}^T\|x_t - \tilde{x}_t \|^2.
\end{equation}
Suppose the offline optimal action sequence is $x_{1:T}^*$, the optimal cost is $\text{cost}(x_{1:T}^*)$. Then we divide both sides of Eqn.~\eqref{eqn:consistency_cost_delay} by $\text{cost}(x_{1:T}^*)$, and get $\forall \lambda_2 >0$,
\begin{equation}\label{eqn:cost_ratio_inequality_3}
\begin{aligned}
    \text{cost}(x_{1:T})  \leq & (1+\lambda_2)\text{cost}(\tilde{x}_{1:T}) + \frac{\beta+ \alpha^2}{2}(1+\frac{1}{\lambda_2}) \cdot\\
    &{\sum_{i=1}^T \biggl( \biggl[\| \tilde{x}_t  - x_t^\pi\| - \sqrt{K \Bigl(d(x_t^\pi, x_{t-p:t-1}^\pi) + \sum_{\tau\in\mathcal{D}_{t}}f(x^{\pi}_{\tau}, y_\tau) \Bigr)}\biggr]^+ \biggr)^2 }
\end{aligned}
\end{equation}

By substituting $K = \frac{2(\lambda - \lambda_0)}{(\beta_h+ \alpha^2) (1 + \frac{1}{\lambda_0})}$ back to Eqn~\eqref{eqn:proof_total_cost}, we have 
\begin{equation}
\begin{aligned}\label{eqn:proof_total_cost}
   \text{cost}(x_{1:T}) \leq& (1+\lambda_2)\text{cost}(\tilde{x}_{1:T}) +(1+\frac{1}{\lambda_2}) \sum_{i=1}^T   \biggl[  \frac{\beta+ \alpha^2}{2} \| \tilde{x}_t  - x_t^\pi\|^2 \\
   &- \frac{\lambda - \lambda_0}{1 + \frac{1}{\lambda_0}} \Bigl(d(x_t^\pi, x_{t-p:t-1}^\pi) + \sum_{\tau\in\mathcal{D}_{t}}f(x^{\pi}_{\tau}, y_\tau) \Bigr)\biggr]^+    
\end{aligned}
\end{equation}
By defining single step cost of the expert $\pi$ as $\text{cost}_t^\pi = d(x_t^\pi, x_{t-p:t-1}^\pi) + \sum_{\tau\in\mathcal{D}_{t}}f(x^{\pi}_{\tau}, y_\tau)$ and the auxiliary cost as $\Delta(\lambda) = \sum_{i=1}^T   \Bigl[   \| \tilde{x}_t  - x_t^\pi\|^2 -  \frac{2(\lambda - \lambda_0)}{(\beta_h+ \alpha^2) (1 + \frac{1}{\lambda_0})}   \text{cost}_t^\pi   \Bigr]^+$
\begin{equation}
    \text{cost}(x_{1:T}) \leq \left(\sqrt{\text{cost}(\tilde{x}_{1:T})} + \sqrt{\frac{\beta+ \alpha^2}{2} \Delta(\lambda)} \right)^2
\end{equation}
Combined with Lemma \ref{lemma:cost_ratio_expert_delay}, we obtain the following bound, which finished this proof.

\begin{equation}
    \text{cost}(x_{1:T}) \leq \min \Biggl((1+\lambda) \text{cost}({x}_{1:T}^\pi), \biggl(\sqrt{\text{cost}(\tilde{x}_{1:T})} + \sqrt{\frac{\beta+ \alpha^2}{2} \Delta(\lambda)} \biggr)^2 \Biggr)
\end{equation}

\subsection{Proof of \Cref{thm:average_non_train}}\label{sec:averagecostpropositionproof}
\begin{proof}
We first give the formal definition of Rademacher complexity of the ML model space with robustification.
\begin{definition}[Rademacher Complexity]\label{definition:rademacher}
	Let $\mathsf{Rob}_{\lambda}(\mathcal{W})=\left\{\mathsf{Rob}_{\lambda}(h_W),W\in\mathcal{W}\right\}$ be the ML model space with robustification constrained by \eqref{eqn:reservation_hitting_delay}. Given the dataset $\mathcal{S}$, the Rademacher complexity with respect to $\mathsf{Rob}_{\lambda}(\mathcal{W})$ is
	\[
\mathrm{Rad}_{\mathcal{S}}(\mathsf{Rob}_{\lambda}(\mathcal{W}))=\frac{1}{|\mathcal{S}|}\mathbb{E}_{\nu}\left[\sup_{W\in\mathcal{W}}\left(\sum_{i\in\mathcal{S}}\nu_i\mathsf{Rob}_{\lambda}\left( h_W(y^i)\right) \right) \right],
	\]
	where $y^i$ is the $i$-th sample in $\mathcal{S}$, and  $\nu_1,\cdots, \nu_n$ are independently drawn from Rademacher distribution.
\end{definition}
Since the cost functions are smooth, they are locally Lipschitz continuous for the bounded action space, and we can apply the generalization bound based on Rademacher complexity \citenewcite{generation_rademacher_bartlett2002rademacher} for the space of robustified ML model $\mathsf{Rob}_{\lambda}(h_W)$. Given any ML model $h_W$ trained on dataset $\mathcal{S}$, with probability at least $1-\delta, \delta\in(0,1)$,
\begin{equation}\label{eqn:generalizationboundproof}
    \mathbb{E}_{\mathbb{P}_y'}[\mathrm{cost}_{1:T}]\leq \overline{\mathrm{cost}}_{\mathcal{S}}(\mathsf{Rob}_{\lambda}(h_{W}))
 +2\Gamma_x \mathrm{Rad}_{\mathcal{S}}(\mathsf{Rob}_{\lambda}(\mathcal{W}))+3\bar{c}\sqrt{\frac{\log(2/\delta)}{|\mathcal{S}|}},
\end{equation}
where $\Gamma_x=\sqrt{T}|\mathcal{X}|\left[\beta_h+\frac{1}{2}(1+\sum_{i=1}^pL_i)(1+\sum_{i=1}^pL_i)\right]$ with $|\mathcal{X}|$ being the size of the action space $\mathcal{X}$ and $\beta_h$, $L_i$, and $p$ as  the smoothness constant, Lipschitz constant of the nonlinear term in the switching cost, and the memory length as defined in Assumptions \ref{assumption:hitting} and \ref{assumption:switching}, and $\bar{c}$ is the upper bound of the total cost for an episode.
We can get the average cost bound in Proposition \ref{thm:average_non_train}.  

Next, we prove that the Rademacher complexity of the ML model space with robustification is no larger than the Rademacher complexity  of the ML model space without robustification expressed as $\{h_W,W\in\mathcal{W}\}$, i.e. we need to prove $\mathrm{Rad}_{\mathcal{S}}(\mathsf{Rob}_{\lambda}(\mathcal{W}))\leq \mathrm{Rad}_{\mathcal{S}}\left(\mathcal{W}\right)$.  The Rademacher complexity can be expressed by Dudley’s entropy integral \citenewcite{Rademacher_complexity_covering_number} as 
\begin{equation}
\mathrm{Rad}_{\mathcal{S}}(\mathsf{Rob}_{\lambda}(\mathcal{W}))=\mathcal{O}\left(\frac{1}{\sqrt{|\mathcal{S}|}}\int_{0}^{\infty}\sqrt{\log\mathbb{N}(\epsilon, \mathsf{Rob}_{\lambda}({\mathcal{W}}),L_2(\mathcal{S}))}\mathrm{d}\epsilon\right),
\end{equation} 
where $\mathbb{N}(\epsilon, \mathsf{Rob}_{\lambda}({\mathcal{W}}), L_2(\mathcal{S}))$ is the covering number \citenewcite{Rademacher_complexity_covering_number} with respect to radius $\epsilon$ and the function distance metric $\|h_1-h_2\|_{L_2(\mathcal{S})}=\frac{1}{|\mathcal{S}|}\sum_{i\in \mathcal{S}}\|h_1(x_i)-h_2(x_i)\|^2$ where $h_1$ and $h_2$ are two functions defined on the space including dataset $\mathcal{S}$. 
 We can find that for any two different weights $W_1$ and $W_2$, their corresponding post-robustification distance $\|\mathsf{Rob}_{\lambda}(h_{W_1})-\mathsf{Rob}_{\lambda}(h_{W_2})\|_{L_2(\mathcal{S})}$ is no larger than their 
pre-robustification distance $\|h_{W_1}-h_{W_2}\|_{L_2(\mathcal{S})}$. To see this, we discuss three cases given any input sample $y$. If both $h_{W_1}(y)$ and $h_{W_2}(y)$ lie in the projection set, then $\mathsf{Rob}_{\lambda}(h_{W_1})(y)=h_{W_1}(y)$ and $\mathsf{Rob}_{\lambda}(h_{W_2})(y)=h_{W_2}(y)$. If $h_{W_1}(y)$ lies in the projection set while $h_{W_2}(y)$ is out of the projection set, the projection operation based on the closed convex projection set will make $\|\mathsf{Rob}_{\lambda}(h_{W_1})(y)-\mathsf{Rob}_{\lambda}(h_{W_2})(y)\|$ to be less than $\|h_{W_1}(y)-h_{W_2}(y)\|$. If both $h_{W_1}(y)$ and $h_{W_2}(y)$ lie out of the projection set, we still have $\|\mathsf{Rob}_{\lambda}(h_{W_1})(y)-\mathsf{Rob}_{\lambda}(h_{W_2})(y)\|\leq \|h_{W_1}(y)-h_{W_2}(y)\|$ since the projection set at each round is a closed convex set \citenewcite{Projection_to_convex_set}. Therefore, after robustification, the distance between two models with different weights will not become larger, i.e. $\|\mathsf{Rob}_{\lambda}(h_{W_1})-\mathsf{Rob}_{\lambda}(h_{W_2})\|_{L_2(\mathcal{S})}\leq \|h_{W_1}-h_{W_2}\|_{L_2(\mathcal{S})}$, which means \ouralg has a covering number $\mathbb{N}(\epsilon, \mathsf{Rob}_{\lambda}({\mathcal{W}}),L_2(\mathcal{S}))$ no larger than that of the individual ML model $\mathbb{N}(\epsilon, {\mathcal{W}},L_2(\mathcal{S}))$ for any $\epsilon$. Thus the Rademacher complexity with the robustification procedure does not increase.

By \citenewcite{rademacher_complexity_neural_network_bartlett2017spectrally}, the upper bound of Rademacher complexity with respect to the space of ML model $\mathrm{Rad}_{\mathcal{S}}(\mathsf{Rob}_{\lambda}(\mathcal{W}))$ is in the order of $\mathcal{O}(\frac{1}{\sqrt{|\mathcal{S}|}})$. Since the Rademacher complexity with the robustification procedure satisfies $\mathrm{Rad}_{\mathcal{S}}(\mathsf{Rob}_{\lambda}(\mathcal{W}))\leq \mathrm{Rad}_{{\mathcal{S}}}\left(\mathcal{W}\right)$, it also decreases with the dataset size in the order of $\mathcal{O}(\frac{1}{\sqrt{|\mathcal{S}|}})$.
\end{proof}

\section{\Train Training}

\Cref{thm:average_non_train} also shows
the benefits of training the ML model in a \train manner.
Specifically, by 
comparing  the losses in ~\eqref{eqn:optimal_w_non_train} and \eqref{eqn:optimal_w_train},
we  see that using \eqref{eqn:optimal_w_train} as
the \train loss for training $W$ can reduce
the term $\overline{\mathrm{cost}}_{\mathcal{S}}(\mathsf{ROB}(h_{W}))$ in
the average cost bound, which
 matches exactly with the training objective in \eqref{eqn:optimal_w_train}. The \train approach is 
only beginning to be explored in the ML-augmented algorithm literature and
non-trivial (e.g., unconstrained downstream optimization in \citenewcite{Shaolei_L2O_ExpertCalibrated_SOCO_SIGMETRICS_Journal_2022_dup}),
especially considering that \eqref{eqn:proj_multi_delay_constraint}
is a constrained optimization problem 
with no explicit gradients.

Gradient-based optimizers such as  
 Adam \citenewcite{DNN_Book_Goodfellow-et-al-2016}
are the de facto state-of-the-art algorithms
for training ML models, offering better optimization results,
convergence, and stability compared to those non-gradient-based alternatives \citenewcite{ML_DifferentiableSimulatorPolicyGradient_ICML_2022_pmlr-v162-suh22b}.
Thus, it is crucial to derive the gradients of the loss function
with respect to the ML model weight $W$ given the added robustification step.

Next, we derive the gradients of $x_t$ with respect to
 $\tilde{x}_t$. For the convenience of presentation, 
 we use the basic SOCO setting with
 a single-step switching cost and no hitting
 cost delay as an example, while noting that the
 same technique can be extended to derive
 gradients in more general settings.
Specifically, for this setting, 
the pre-robustification prediction is 
given by $\tilde{x}_t = h_W(\tilde{x}_{t-1}, y_t)$, where $W$ denotes the ML model
weight. Then, the actual post-robustification action $x_t$ is obtained by projection in \eqref{eqn:proj_multi_delay_constraint}
by setting $q=0$ and $p=1$, given the ML prediction $\tilde{x}_t$, the expert's action  $x_{t}^{\pi}$ and cumulative  $\mathrm{cost}(x_{1:t}^{\pi})$ up to $t$, and the actual
cumulative  $\text{cost}(x_{1:t-1})$ up to $t-1$.

The gradient of
the loss function $\mathrm{cost}(x_{1:T})=\sum_{t=1}^T\left(f(x_t, y_t) + d(x_t, x_{t-1}) \right)$ with respect to the ML model weight $W$ is given by 
$\sum_{t=1}^T\nabla_W \big( f(x_t, y_t) + d(x_t, x_{t-1}) \big)$.
Next, we write the gradient of per-step cost with with respect
to $W$ as follows:
\begin{equation}
    \begin{split}\label{eqn:projection_chain_rule} 
    &\nabla_W \big( f(x_t, y_t) + d(x_t, x_{t-1}) \big) \\
    = & \nabla_{x_t} \big( f(x_t, y_t) + d(x_t, x_{t-1}) \big) \nabla_W x_t + \nabla_{x_{t-1}} \big( f(x_t, y_t) + d(x_t, x_{t-1}) \big) \nabla_W x_{t-1}\\
    =&\nabla_{x_t} \big( f(x_t, y_t) + d(x_t, x_{t-1}) \big) 
    \nabla_W x_t + \nabla_{x_{t-1}}  d(x_t, x_{t-1})  \nabla_W x_{t-1},
    \end{split}
\end{equation}
where the gradients $\nabla_{x_t} \big( f(x_t, y_t) + d(x_t, x_{t-1}) \big)$ 
and $\nabla_{x_{t-1}}  d(x_t, x_{t-1})$
are trivial given the hitting and switching cost functions,
and the gradient $\nabla_W x_{t-1}$  is obtained 
at time $t-1$ in the same way as 
$\nabla_W x_t$.
To derive $\nabla_W x_t$, by the chain rule,
we have: 
\begin{equation}\label{eqn:gradient_x_t_w}
 \nabla_W x_t=\nabla_{\tilde{x}_t} x_t \nabla_W\tilde{x}_t %h_W(\tilde{x}_{t-1},y_t) 
 + \nabla_{\text{cost}(x_{1:t-1})} x_t \nabla_W \text{cost}(x_{1:t-1}),   
\end{equation}
where
$\nabla_W\tilde{x}_t$ is 
the gradient of the ML output (following a recurrent architecture
illustrated in Fig.~\ref{fig:illustration} in the appendix) with respect to the weight $W$
and can  be obtained recursively by
using  off-the-shelf BPTT optimizers \citenewcite{DNN_Book_Goodfellow-et-al-2016},  
and $\nabla_W \text{cost}(x_{1:t-1}) = \sum_{\tau=1}^{t-1}\nabla_W \big( f(x_\tau, y_\tau) + d(x_\tau, x_{\tau-1}) \big)$ can also be recursively calculated once we have the gradient in Eqn.~\eqref{eqn:projection_chain_rule}.
Nonetheless, it is non-trivial to calculate
the two gradient terms in Eqn.~\eqref{eqn:gradient_x_t_w}, i.e.,
$\nabla_{\tilde{x}_t} x_t$ and $\nabla_{\text{cost}(x_{1:t-1})} x_t$,
where $x_t$ 
itself is the solution to the constrained optimization problem \eqref{eqn:proj_multi_delay_constraint} unlike
in the simpler unconstrained case \citenewcite{Shaolei_L2O_ExpertCalibrated_SOCO_SIGMETRICS_Journal_2022_dup}. 
As we cannot explicitly write $x_t$ 
 in a closed form in terms of $\tilde{x}_t$ and $\text{cost}(x_{1:t-1})$, we leverage
the KKT conditions \citenewcite{BoydVandenberghe,L2O_Survey_Amortized_Continuous_Brandon_arXiv_2022_DBLP:journals/corr/abs-2202-00665_dup,L2O_DifferentiableConvexOptimization_Brandon_NEURIPS2019_9ce3c52f} 
 to implicitly derive $\nabla_{\tilde{x}_t} x_t$ and $\nabla_{\text{cost}(x_{1:t-1})} x_t$  in the next proposition.

\begin{proposition}[Gradients by KKT conditions]\label{thm:back-propogation}
Let $x_t\in\mathcal{X}$ and $\mu\geq0$ be the primal and dual solutions to the problem \eqref{eqn:proj_multi_delay_constraint}, respectively. The gradients of
 $x_t$ 
 with respect to $\tilde{x}_t$ and $\text{cost}(x_{1:t-1})$ are
$$\nabla_{\tilde{x}_t} x_t =  \Delta_{11}^{-1} [I + \Delta_{12} Sc(\Delta,\Delta_{11})^{-1}\Delta_{21} \Delta_{11}^{-1} ],$$
$$\nabla_{\text{cost}(x_{1:t-1})} x_t = \Delta_{11}^{-1} \Delta_{12} Sc(\Delta, \Delta_{11})^{-1} \mu,$$
where $\Delta_{11} = I + \mu \Big( \nabla_{x_t,x_t}  f(x_t, y_t) + \left(1 + (1+\frac{1}{\lambda_0}) (L_1^2 + L_1)\right) I  \Big)$, $\Delta_{12} = \nabla_{x_t} f(x_t, y_t) + \left( x_t - \delta(x_{t-1}) \right)  +  \left(1 + (1+\frac{1}{\lambda_0}) (L_1^2 + L_1)\right) (x_t -  x_t^\pi)$, $\Delta_{21} = \mu \Delta_{12}^\top$, $\Delta_{22} = f(x_t, y_t) + d(x_t, x_{t-1}) + G(x_t,x_t^\pi) + \text{cost}(x_{1:t-1}) -  (1+ \lambda) \text{cost}(x_{1:t}^\pi )$, and $Sc(\Delta, \Delta_{11})=\Delta_{22}-\Delta_{21}\Delta^{-1}_{11}\Delta_{12}$ is the Schur-complement of $\Delta_{11}$ in the blocked matrix $\Delta = \big[[\Delta_{11}, \Delta_{12}],[\Delta_{21}, \Delta_{22}] \big]$.   
\end{proposition}

If the ML prediction $\tilde{x}_t$ happens to lie on the 
boundary such that the inequality in \eqref{eqn:proj_multi_delay_constraint}
becomes an equality for $x=\tilde{x}_t$, 
then the gradient does not exist in this case
and $Sc(\Delta, \Delta_{11})$ may not be full-rank. 
Nonetheless, we can still calculate the pseudo-inverse 
of $Sc(\Delta, \Delta_{11})$ 
and use  \Cref{thm:back-propogation} to 
calculate the subgradient. Such approximation is actually
a common practice to address non-differentiable points
for training ML models, e.g., 
using $0$ as the subgradient
of $ReLu(\cdot)$ at the 
zero point  \citenewcite{DNN_Book_Goodfellow-et-al-2016}.

% Add separate reference for appendix
\clearpage
\bibliographystylenewcite{unsrt}
\bibliographynewcite{newcite.bbl}
%\bibliographynewcite{ref_jy_knowledge,ref_pengfei,ref_ren, ref_duplicate}

\end{document}